# Parsing English with a Link Grammar

Daniel D. K. Sleator *     Davy Temperley †

October 1991

CMU-CS-91-196

School of Computer Science
Carnegie Mellon University
Pittsburgh, PA 15213

## Abstract

We develop a formal grammatical system called a *link grammar*, show how English grammar can be encoded in such a system, and give algorithms for efficiently parsing with a link grammar. Although the expressive power of link grammars is equivalent to that of context free grammars, encoding natural language grammars appears to be much easier with the new system. We have written a program for general link parsing and written a link grammar for the English language. The performance of this preliminary system – both in the breadth of English phenomena that it captures and in the computational resources used – indicates that the approach may have practical uses as well as linguistic significance. Our program is written in C and may be obtained through the internet.

* School of Computer Science, Carnegie Mellon University, Pittsburgh, PA 15213, `sleator@cs.cmu.edu`.

† Music Department, Columbia University, New York, NY 10027, `dt3@cunixa.cc.columbia.edu`.

Research supported in part by the National Science Foundation under grant CCR-8658139, Olin Corporation, and R. R. Donnelley and Sons.

The views and conclusions contained in this document are those of the authors and should not be interpreted as representing the official policies, either expressed or implied, of Olin Corporation, R. R. Donnelley and Sons, or the NSF.

# 1. Introduction

Most sentences of most natural languages have the property that if arcs are drawn connecting each pair of words that relate to each other, then the arcs will not cross [8, p. 36]. This well-known phenomenon, which we call *planarity*, is the basis of *link grammars* our new formal language system.

A link grammar consists of a set of *words* (the terminal symbols of the grammar), each of which has a *linking requirement*. A sequence of words is a *sentence* of the language defined by the grammar if there exists a way to draw arcs (which we shall hereafter call *links*) among the words so as to satisfy the following conditions:

Planarity: The links do not cross (when drawn above the words).

Connectivity: The links suffice to connect all the words of the sequence together.

Satisfaction: The links satisfy the linking requirements of each word in the sequence.

The linking requirements of each word are contained in a *dictionary*. To illustrate the linking requirements, the following diagram shows a simple dictionary for the words *a*, *the*, *cat*, *snake*, *Mary*, *ran*, and *chased*. The linking requirement of each word is represented by the diagram above the word.

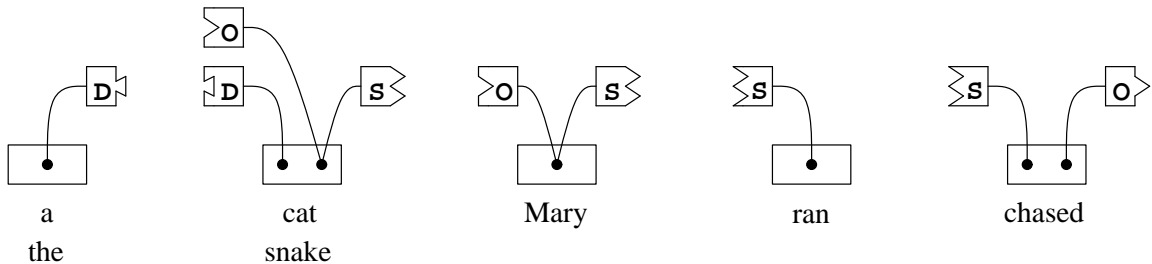

Each of the intricately shaped labeled boxes is a *connector*. A connector is satisfied by "plugging it into" a compatible connector (as indicated by its shape). If the mating end of a connector is drawn facing to the right, then its mate must be to its right facing to the left. Exactly one of the connectors attached to a given black dot must be satisfied (the others, if any, must not be used). Thus, *cat* requires a D connector to its left, and either an O connector to its left or a S connector to its right. Plugging a pair of connectors together corresponds to drawing a link between that pair of words.

The following diagram shows how the linking requirements are satisfied in the sentence *The cat chased a snake*.

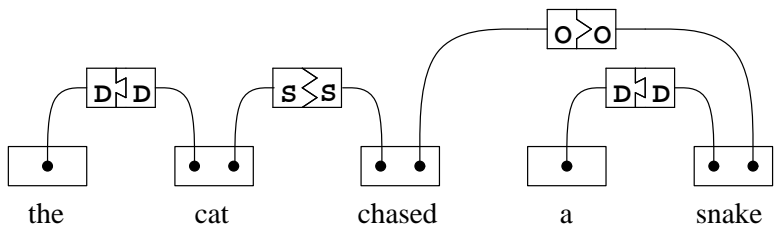

(The unused connectors have been suppressed here.) It is easy to see that *Mary chased the cat*, and *the cat ran* are also sentences of this grammar. The sequence of words: *the Mary chased cat* is not in this language. Any attempt to satisfy the linking requirements leads to a violation of one of the three rules. Here is one attempt.

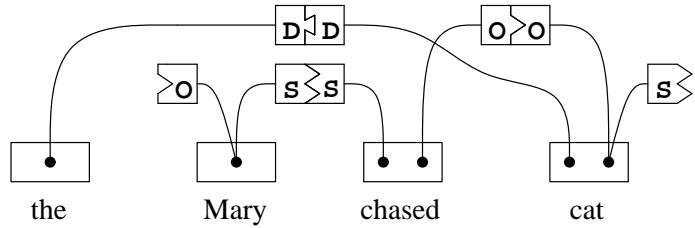

Similarly *ran Mary*, and *cat ran chased* are not part of this language.

A set of links that prove that a sequence of words is in the language of a link grammar is called a *linkage*. From now on we'll use simpler diagrams to illustrate linkages. Here is the simplified form of the diagram showing that *the cat chased a snake* is part of this language:

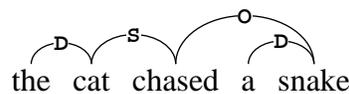

We have a succinct, computer-readable notation for expressing the dictionary of linking requirements. The following dictionary encodes the linking requirements of the previous example.

| words | formula |
|---|---|
| a the | D+ |
| snake cat | D- & (O- or S+) |
| Mary | O- or S+ |
| ran | S- |
| chased | S- & O+ |

The linking requirement for each word is expressed as a formula involving the operators `&`, and `or`, parentheses, and connector names. The `+` or `-` suffix on a connector name indicates the direction (relative to the word being defined) in which the matching connector (if any) must lie. The `&` of two formulas is satisfied by satisfying both the formulas. The `or` of two formulas requires that exactly one of its formulas be satisfied. The order of the arguments of an `&` operator is significant. The farther left a connector is in the expression, the nearer the word to which it connects must be. Thus, when using *cat* as an object, its determiner (to which it is connected with its `D-` connector) must be closer than the verb (to which it is connected with its `O-` connector).

The following dictionary illustrates a slightly more complicated link grammar. (The notation "{*exp*}" means the expression *exp* is optional, and "@A-" means one or more connectors of type `A` may be attached here.)

| words | formula |
|---|---|
| a the | D+ |
| snake cat | {@A-} & D- & {B+} & (O- or S+) |
| chased bit | S- & (O+ or B-) |
| ran | S- |
| big green black | A+ |
| Mary | O- or S+ |

The sentence *the big snake the black cat chased bit Mary* is in the language defined by this grammar because of the following links:

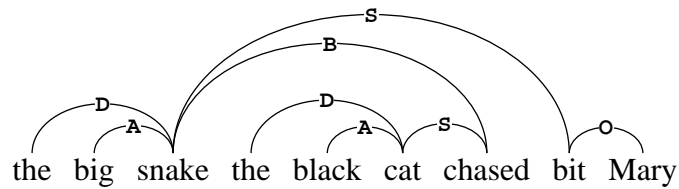

In this case (as in the one above) there is a unique linkage satisfying all the requirements.

We have written a link grammar of roughly seven hundred definitions that captures many phenomena of English grammar. It handles: noun-verb agreement, questions, imperatives, complex and irregular verbs (*wanted*, *go*, *denied*, etc.), different types of nouns (mass nouns, those that take *to*-phrases, etc.), past- or present-participles in noun phrases, commas, a variety of adjective types, prepositions, adverbs, relative clauses, possessives, and many other things.

We have also written a program for parsing with link grammars. The program reads in a dictionary (in a form very similar to the tables above) and will parse sentences (typically typed by the user) according to the given grammar. The program does an exhaustive search – it finds every way of parsing the given sequence with the given link grammar. It is based on our own $O(n^3)$ algorithm ($n$ is the number of words in the sequence to be parsed). The program also makes use of several very effective data structures and heuristics to speed up parsing. The program is comfortably fast – parsing typical newspaper sentences in a few seconds.

Both our program (written in ANSI-C) and our dictionary are available through the internet.[1] Having the program available for experimentation may make it easier to understand this paper.

An interesting aspect of our system is that it has no explicit notion of *constituents* or *categories*.[2]

---

[1] To get the system, use the `ftp` command to connect to host `spade.pc.cs.cmu.edu`. Login with `anonymous` as your user name, and get all the files from the directory `/usr/sleator/public/grammar`. Read the instructions in the file called `README`. That directory contains a version compatible with this paper. When we refer to the *on-line* dictionary, we mean the one in this directory.

[2] In most sentences parsed with our dictionaries, constituents can be seen to emerge as contiguous connected collections of words attached to the rest of the sentence by a particular type of link. For example in the dictionary above, S links always attach a noun phrase (the connected collection of words at the left end of the link) to a verb (on the right end of the link). O links work in a similar fashion. In these cases the links of a sentence can be viewed as an alternative way of specifying the constituent structure of the sentence. On the other hand, this is not the way we think about link grammars, and we see no advantage in taking that perspective.

Constituents are widely regarded as an essential component of any serious linguistic system. Explaining why these are "necessary", Radford [10, p. 53] writes:

> Thus far, we have seen two advantages of category-based grammars: firstly, they capture distributional regularities and, secondly, they can handle recursion. A third advantage is that category-based grammars are more *constrained* than word-based grammars.

Our system is word-based – categories (*i.e.* noun phrase, verb phrase, etc.) play no role – yet it captures distributional regularities, handles recursion, and is sufficiently constrained.

We are not the first to question the "necessity" of constituents for doing syntactic analysis. Hudson [6, Chapter 3] and Mel'čuk [8] argue against it. Instead they advocate the use of *dependency structure* for syntactic and semantic analysis. This structure, as defined by Mel'čuk, consists (in part) of a set of planar directed arcs among the words that form a tree. Each word (except the *root word*) has an arc out to exactly one other word, and no arc may pass over the root word. In a linkage (as opposed to a dependency structure) the links are undirected, and may form cycles, and there is no notion of a root word. Our system can perhaps be viewed as a technique for parsing using the assumption that there exists an underlying dependency structure. We are not aware of any other attempt to do this.[3]

**The organization of this paper**

In section 2 we define link grammars more formally and explain the notation and terminology used throughout the rest of the paper. In section 3 we describe the workings of an abridged version of our large on-line dictionary. This abridged dictionary is reproduced fully in appendix D. Our $O(n^3)$ algorithm is described in section 4, and the data structures and heuristics that make it run fast are described in section 5.

Coordinating conjunctions such as *and* pose a problem for link grammars. This is because in a sentence like *Danny and Davy wrote and debugged a program* there should logically be links between both *Danny* and *Davy* and both *wrote* and *debugged*. Such links would cross. We have

---

[3]Another grammatical system, known as a combinatory categorial grammar (CCG) bears some superficial resemblance to our system. In a CCG, as in our system, each word of the dictionary is given a definition which defines how it can be used in a sentence. A derivation in this system is a sequence of combination operations applied to neighboring words (or groups of words). The operation creates a new, larger group of words, and gives it a definition. The combination operation cancels out one of the symbols in each of the words (or groups) definitions.

Because of the similarities between CCGs and link grammars, we attempted to find a precise connection between them, and failed. To illustrate the difficulties, we explain one approach. A derivation in a CCG can be transformed into a way of drawing links among the words so that they do not cross: a link is drawn between the symbols cancelled by the combination operation. This connection breaks down under more careful scrutiny. In all of the examples of CCGs that we have seen [1, 4, 7, 11], the linkages that result from analyzing English sentences are very different from those coming from our system. For example, Joshi [7] shows a CCG capable of analyzing the sentence: *Harry likes peanuts passionately*. In our interpretation of that analysis, *Harry* ends up being linked to *passionately*. Our system links *Harry* to *likes*.

It thus appears that the way a CCG writer and a link grammar writer think about language when writing their grammars is entirely different. Of course, we can't rule out the possibility of there being a different interpretation of a CCG as a link grammar, or vice versa.

devised a scheme that handles the vast majority of uses of such conjunctions and incorporated it into our program. It is described in section 6.

Certain constructs are difficult to handle only using the basic link grammar framework. One example is the non-referential use of *it*: *It is likely that John will go* is correct, but *The cat is likely that John will go* is wrong. It is possible – but awkward – to distinguish between these with a link grammar. To deal with this (and a number of other phenomena), our system has a *post-processor* which begins with a linkage, analyzes its structure, and determines if certain conditions are satisfied. This allows the system to correctly judge a number of subtle distinctions (including that mentioned above). It is described in section 7.

There are a number of common English phenomena that are not handled by our current system. In section 8 we describe some of these, along with reasons for this state of affairs. The reasons range from the fact that ours is a preliminary system to the fact that some phenomena simply do not fit well into the link grammar framework.

Having pointed out what our system can't do, we felt justified to include a section showing what it can do. In appendix A, we offer a long list of sentences which illustrate some of the more advanced features of the on-line dictionary and post-processing system.

We have tested and tuned our system by applying it to many articles taken from newspapers. Appendix B contains a transcript of our program analyzing two articles taken from the New York Times. The linkage diagrams for these sentences are displayed there. The amount of computation time required for our program to do each part of the processing of each of the sentences of the transcript is in appendix C.

Ignoring post-processing, all variants of link grammars considered in this paper are context free. Appendix E contains a proof of this fact.

Section 9 contains a few concluding remarks about the advantages of our system, and the role our ideas might play in several natural language applications.

**Contents**

## 2. Definitions and notation

In this section we define several variants of link grammars, and introduce the notation used throughout the rest of the paper.

### 2.1. Basic link grammars

The link grammar dictionary consists of a collection of entries, each of which defines the linking requirements of one or more words. These requirements are specified by means of a *formula* of *connectors* combined by the binary associative operators `&` and `or`. Presidence is specified by means of parentheses. Without loss of generality we may assume that a connector is simply a character string ending in `+` or `-`.

When a link connects to a word, it is associated with one of the connectors of the formula of that word, and it is said to *satisfy* that connector. No two links may satisfy the same connector. The connectors at opposite ends of a link must have names that *match*, and the one on the left must end in `+` and the one on the right must end in `-`. In basic link grammars, two connectors match if and only if their strings are the same (up to but not including the final `+` or `-`). A more general form of matching will be introduced later.

The connectors satisfied by the links must serve to satisfy the whole formula. We define the notion of satisfying a formula recursively. To satisfy the `&` of two formulas, both formulas must be satisfied. To satisfy the `or` of two formulas, one of the formulas must be satisfied, and *no* connectors of the other formula may be satisfied. It is sometimes convenient to use the empty formula ("`()`"), which is satisfied by being connected to no links.

A sequence of words is a *sentence* of the language defined by the grammar if there exists a way to draw links among the words so as to satisfy each word's formula, and the following *meta-rules*:

Planarity:   The links are drawn above the sentence and do not cross.

Connectivity:   The links suffice to connect all the words of the sequence together.

Ordering:   When the connectors of a formula are traversed from left to right, the words to which they connect proceed from near to far. In other words, consider a word, and consider two links connecting that word to words to its left. The link connecting the nearer word (the shorter link) must satisfy a connector appearing to the left (in the formula) of that of the other word. Similarly, a link to the right must satisfy a connector to the left (in the formula) of a longer link to the right.

Exclusion:   No two links may connect the same pair of words.

## 2.2. Disjunctive form

The use of formulas to specify a link grammar dictionary is convenient for creating natural language grammars, but it is cumbersome for mathematical analysis of link grammars, and in describing algorithms for parsing link grammars. We therefore introduce a different way of expressing a link grammar called *disjunctive form*.

In disjunctive form, each word of the grammar has a set of *disjuncts* associated with it. Each disjunct corresponds to one particular way of satisfying the requirements of a word. A disjunct consists of two ordered lists of connector names: the *left list* and the *right list*. The left list contains connectors that connect to the left of the current word (those connectors end in -), and the right list contains connectors that connect to the right of the current word. A disjunct will be denoted:

$$((L_1, L_2, \ldots, L_m) \ (R_n, R_{n-1}, \ldots, R_1))$$

Where $L_1, L_2, \ldots, L_m$ are the connectors that must connect to the left, and $R_1, R_2, \ldots, R_n$ are connectors that must connect to the right. The number of connectors in either list may be zero. The trailing + or - may be omitted from the connector names when using disjunctive form, since the direction is implicit in the form of the disjunct.

To satisfy the linking requirements of a word, one of its disjuncts must be satisfied (and no links may attach to any other disjunct). To satisfy a disjunct all of its connectors must be satisfied by appropriate links. The words to which $L_1, L_2, \ldots$ are linked are to the left of the current word, and are monotonically increasing in distance from the current word. The words to which $R_1, R_2, \ldots$ are linked are to the right of the current word, and are monotonically increasing in distance from the current word.

It is easy to see how to translate a link grammar in disjunctive form to one in standard form. This can be done simply by rewriting each disjunct as

$$( L_1 \ \& \ L_2 \ \& \ \cdots \ \& \ L_m \ \& \ R_1 \ \& \ R_2 \ \& \ \cdots \ \& \ R_n \ )$$

and combining all the disjuncts together with the `or` operator to make an appropriate formula.

It is also easy to translate a formula into a set of disjuncts. This is done by enumerating all ways that the formula can be satisfied. For example, the formula (adapted from the introduction):

$$\text{(A- or ())} \ \& \ \text{D-} \ \& \ \text{(B+ or ())} \ \& \ \text{(O- or S+)}$$

corresponds to the following eight disjuncts:

```
     ((A,D)   (S,B))
    ((A,D,O)  (B))
     ((A,D)   (S))
    ((A,D,O)  ())
       ((D)   (S,B))
      ((D,O)  (B))
       ((D)   (S))
      ((D,O)  ())
```

## 2.3. Our dictionary language

To streamline the difficult process of writing the dictionary, we have incorporated several other features to the dictionary language. Examples of all of these features can be found in abundance in the abridged dictionary of appendix D, and in section 3.

It is useful to consider connector matching rules that are more powerful than simply requiring the strings of the connectors to be identical. The most general matching rule is simply a table – part of the link grammar – that specifies all pairs of connectors that match. The resulting link grammar is still context-free. (See appendix E.)

In the dictionary presented later in this paper, and in our larger on-line dictionary, we use a matching rule that is slightly more sophisticated than simple string matching. We shall now describe this rule.

A connector name begins with one or more upper case letters followed by a sequence of lower case letters or *s. Each lower case letter (or *) is a *subscript*. To determine if two connectors match, delete the trailing + or -, and append an infinite sequence of *s to both connectors. The connectors match if and only if these two strings match under the proviso that * matches a lower case letter (or *).

For example, S matches both Sp and Ss, but Sp does not match Ss. Similarly, D*u, matches Dmu and Dm, but not Dmc. All four of these connectors match Dm. (These examples are taken from section 3.)

The formula "((A- & B+) or ())" is satisfied either by using both A- and B+, or by using neither of them. Conceptually, then, the the expression "(A+ & B+)" is optional. Since this occurs frequently, we denote it with curly braces, as follows: {A+ & B+}.

It is useful to allow certain connectors to be able to connect to one or more links. This makes it easy, for example, to allow any number of adjectives to attach to a noun. We denote this by putting an "@" before the connector name, and call the result a *multi-connector*. Context-freeness is unaffected by this change, as shown in appendix E.

Our dictionaries consist of a sequence of *entries*, each of which is a list of words separated by spaces, followed by a colon, followed by the formula defining the words, followed by a semicolon.

If a word (such as *move* or *can*) has more than one distinct meaning, it is useful to be able to give it two definitions. This is accomplished by defining several versions of the word with differing suffixes. The suffix always begins with a ".", which is followed by one or more characters. We use the convention that ".v" indicates a verb and ".n" indicates a noun (among others). When the user types the "move", the program uses an expression that is equivalent to that obtained by `oring` the expressions for the two versions of the word. When it prints out the linkage, it uses whichever version is appropriate for that particular linkage.

As of this writing there is no macro facility in the dictionary language. There is reason to believe that using macros would significantly reduce the size of the dictionary while making it easier to understand.

### 2.4. Alternative meta-rules

The ordering requirement can be considered a property of the `&` operator (since at most one operand of an `or` will have any satisfied links at all). An `&`-like operator which is immune to such an ordering constraint might be very useful for creating a link grammar for a language in which word order is not as constrained as it is in English. Such a generalized link grammar can be expressed in disjunctive form, so the proof of context-freeness and the algorithm of section 4 are valid. The number of disjuncts will, however, increase.

It is natural to ask what the effect of modifying the other meta-rules might be. For example what happens if you eliminate the exclusion rule? What about restricting the linkages to be acyclic? We have given very little thought to alternative meta-rules, and to the effect they would have on the dictionary, the parsing algorithms, and post-processing. This is a topic for future research.

## 3. The workings of a link grammar for English

In this section we illustrate the use of link grammars through a very detailed example of a link grammar for English. The dictionary we describe is reproduced fully in appendix D[4]. Most of the expressions below are not exactly as they appear in the appendix; our description begins gently with very basic features, and gradually builds up the expressions as we explain more and more features. This description models the way in which the dictionary might evolve as it is being written.

The on-line dictionary has many features and capabilities which are not discussed here. Some examples of these features are shown in section A.

### 3.1. Simple things about verbs and nouns (S, I, GI, V, T)

All nouns have an S+ connector. All simple verbs and auxiliaries have an S- connector. So:

>   dog bird idea people week: S+;
>   run runs ran hits gives expects: S-;

Nominative pronouns also have an S+ connector:

>   he she I: S+;

Words are subscripted to insure noun-verb agreement:

>   dog idea week he it: Ss+;
>   dogs ideas weeks they we: Sp+;
>
>   runs kicks expects: Ss-;
>   run kick expect: Sp-;

Simple-past forms of verbs simply have an un-subscripted S-, meaning they can take either singular or plural:

>   ran kicked expected: S-;

Infinitive forms of verbs have an I-. Most infinitive verb forms are the same as the simple plural form, so a single dictionary entry serves both of these functions:

>   kick give expect: S- or I-;

Verbs which take infinitives - *do*, *may*, *can*, etc. - have an I+ connector. They also have an S- connector. The two connectors are &ed; both must be satisfied for the word's use to be valid.

>   may can: S- & I+;

Past participle forms of verbs have a T- connector. These forms are usually the same as simple past forms, so again one entry usually serves for both:

---
[4]It is also in /usr/sleator/public/grammar/abridged-dictionary

kicked expected: `S-` or `T-`;

Forms of the verb *have* have a `T+` connector, &ed with their `S-` connectors.

Present participles of verbs – *ing* words – have a `GI-` connector:

running giving expecting: `GI-`;

Verbs that take *-ing* words, like *be*, *like*, and *enjoy*, have `GI+` connectors.

Passive forms of verbs – which are usually the same as past participles – have a `V-` connector. Forms of the verb *be* have a `V+` connector.

Irregular verbs can be dealt with under this system quite easily. For example, consider the verb *hit*. There, the present-plural form, simple-past form, infinitive, and past participle are all the same. Therefore they are all included under a single dictionary entry:

hit: `S-` or `I-` or `T-`;

### 3.2. Question-inversion (`SI`)

Verbs that can be inverted with questions have `SI+` connectors as well as `S-` connectors. Nouns have `SI-` connectors as well as `S+` connectors.

dog cat he: `S+` or `SI-`;
does has: `Ss-` or `SIs+`;
can may: `S-` or `SI+`;

Combining the features discussed above, consider the entry for *has*:

has: (`Ss-` or `SIs+`) & `T+`;

Here, for the first time, the ordering of elements in the expression is important. The fact that the `T+` element comes after the `SIs+` element means that the `SIs` connection has to be made closer than the `T` connection. In other words, this rule allows

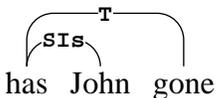

and prevents

*Has gone John

### 3.3. Determiners (`D`)

Nouns have a `D-` connector and determiners have a `D+` connector. On singular nouns, the `D-` is mandatory; on plural nouns and mass nouns, it is optional. Some nouns, like *war*, can be either singular or mass; here the `D-` is optional.

```
dog: D- & (Ss+ or SIs-);
dogs: {D-} & (Sp+ or SIp-);
war: {D-} & (Ss+ or SIs-);
wars: {D-} & (Sp+ or SIp-);
decency: {D-} & (Sp+ or SIp-);
decencies: * (not in the dictionary)
```

Some determiners can be used with any noun, like *the* and *his*. Others can be only used with singular nouns, like *a*. Others can only be used with mass or plural nouns, like *all* or *most*. Others, like *many*, can only be used with plural nouns. We use subscripts to enforce these rules:

```
the: D+;
a: Ds+;
many: Dmc+;
all: Dm+;

dog: Ds- & (Ss+ or SIs-);
dogs: {Dmc-} & (Sp+ or SIp-);
war: {D*u-} & (Ss+ or SIs-);
wars: {Dmc-} & (Sp+ or SIp-);
decency: {Dmu-} & (Sp+ or SIp-);
```

### 3.4. Transitive verbs (`O`)

Transitive verbs have an `O+` connector. Nouns have an `O-` connector.

```
dog idea week: Ds- & (Ss+ or SIs- or O-);
chases kills admonishes: Ss- & O+;
killed admonished: (S- or T-) & O+;
chase kick: (S- or I-) & O+;
kicking: GI- & O+;
```

Accusative pronouns have an `O-` connector, no `S+` connector.

```
him me: O-;
```

Note that on nouns, the `O-` connector is ored with the `S+` and `SI-` connectors. This means, essentially, that a noun can serve as a subject or an object, but not both. (We'll get to relative clauses shortly.) On verbs, however, the `O+` is &ed with the `S-`. This means that a transitive verb must have both a subject and an object for its use to be valid.

Some verbs are optionally transitive; they may take an object, or they may not. Such verbs have optional `O+` connectors:

```
moves drives: Ss- & {O+};
```

How about passive? We mentioned that passive forms of verbs have a `V-` connector. Naturally, only transitive verbs have `V-` connectors. However, the `V-` connector can not simply be ored with the `T-` (past participle) connector; this would allow

*John was kicked the dog

Rather, it has to be ored with everything:

kicked: ((S- or T-) & O+) or V-;

This, then, will correctly judge the following:

John kicked the dog
John has kicked the dog
*John has kicked
John was kicked
*John was kicked the dog

### 3.5. More complex verbs (`TO`, `TH`, `CL`, `R`, `AI`)

Some verbs take two objects, such as *give*. Such verbs have two `O` connectors.

give: (S- or I-) & O+ & {O+};

(We consider the second of these connectors to be optional, so that *John gave a speech* is a valid sentence.)

Some verbs take infinitives with *to*. These verbs have a `TO+` connector. The word *to* has a `TO-` connector and an `I+` connector.

threaten demand: (Ss- or I-) & TO+;
to: TO- & I+;

This therefore allows sentences like *I want to go to the store*, *John expects to be appointed*.

Other verbs take objects *and* infinitives. These verbs carry both an `O+` and a `TO+`, although the `TO+` is usually optional.

urges designs causes allows encourages: Ss- & O+ & {TO+};

Some verbs take *that* plus a clause, or just a clause with *that* being unnecessary. For these, `TH+` and `CL+` connectors are used. Verbs that take clauses have `CL+` connectors; complete clauses have `CL-` connectors. (This will be explained shortly.) Verbs that take *that* plus a clause have a `TH+`; *that* has a `TH-` and a `CL+`.

protests complains: Ss- & TH+;
assumes supposes: Ss- & (TH+ or CL+);
that: TH- & CL+;

Some verbs take question words, like *who*, *what*, and *when*. Such verbs have an `R+` connector. Question words have an `R-` connector.

wonders: Ss- & R+;

Some verbs take adjectives. These verbs have an `AI+` connector. Adjectives have an `AI-` connector (more about adjectives below).

looks feels: `Ss- & AI+`;

Some verbs take the subjunctive: a nominative noun-phrase, plus an infinitive. This can be expressed using the connectors we've already discussed:

demands: `Ss- & (SI+ & I+)`;

Other verbs combine these features in a wide variety of ways. The entries for these verbs comprise most of the dictionary. To give just a couple of examples: the verb *see* can take just an object, or an object plus an infinitive (without *to*), or an object plus a present participle:

John sees the bird
John sees the bird fly
John sees the bird flying
*John sees fly
*John sees fly the bird

So the dictionary entry is as follows:

sees: `S- & O+ & {I+ or GI+}`;

As another example, consider the verb *expect*. This can take an object; an object plus a *to-infinitive* phrase; a *to-infinitive* phrase alone; a *that + clause* phrase; or just a clause.

John expected the party
John expects me to go
John expects to go
John expects that I will go
John expects I will go

So the entry is as follows:

expects: `Ss- & (CL+ or TH+ or TO+ or (O+ & {TO+}))`;

Each form of a verb has its own definition. In the case of *expect*, then, the dictionary contains the following:

expect: `(Sp- or I-) & (CL+ or TH+ or TO+ or (O+ & {TO+}))`;
expects: `Ss- & (CL+ or TH+ or TO+ or (O+ & {TO+}))`;
expected: `(S- or T-) & (CL+ or TH+ or TO+ or (O+ & {TO+}))`;
expecting: `GI- & (CL+ or TH+ or TO+ or (O+ & {TO+}))`;

Again, passive forms of verbs take `V-` connectors, but they must be ored with the other connectors so that they can only be used when the verb is serving a transitive function. In the case of *expected*, then:

expected: `((S- or T-) & (CL+ or TH+ or TO+ or (O+ & {TO+})))
or (V- & {TO+})`;

This, then, will correctly judge the following:

John expected to go
John expected Joe to go
John expected Joe
*John expected
John was expected to go
*John was expected Joe to go
*John was expected that Joe would go
John expected that Joe would go
*John expected Joe that Sam would go

### 3.6. Adjectives (`A`, `AI`)

Adjectives have an `A+`; nouns have an `@A-`. Recall that the `@` allows the connector to link to one or more other connectors, thus allowing

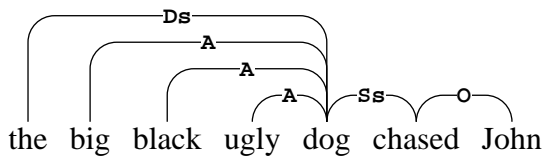
the big black ugly dog chased John

We put the `@A-` first on a noun, because the adjective must be closer than anything else, and we surround it by curly braces to make it optional.

   big black ugly: `A+`;
   dog: `{@A-} & Ds- & (Ss+ or SIs- or O-)`;

Adjectives also have an `AI-` connector; verbs that take adjectives, such as *be* and *seem*, have an `AI+`:

   is seems: `Ss- & AI+`;
   big black ugly: `A- or AI+`;

Adjectives have many of the same complexities as verbs: they can take *to-* phrases, *that-* phrases, and question-word phrases. These things are handled with the same connectors used on verbs.

   eager reluctant able: `A- or (AI+ & {TO+})`;
   confident doubtful aware: `A- or (AI+ & {TH+})`;
   certain: `A- or (AI+ & {TO+ or TH+ or R+})`;

Notice that the `TO+` and `TH+` connectors on adjectives are &ed with the `AI-`, but ored with the `A+`; that is, these connectors can be used only when the adjective follows the subject, not when it precedes the subject.

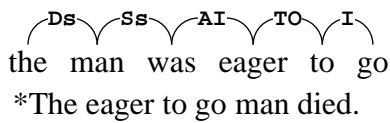
the man was eager to go
   *The eager to go man died.

Note also that the `TO+` and `TH+` connections on adjectives are always optional. In our opinion, it is always grammatical to simply use such an adjective without a *to* or *that* phrase.

Some adjectives – and a few verbs – can only take *to-* and *that-* phrases when the subject is *it*. For example, *It is likely that John will come* is correct, whereas *Mary is likely that John will come* is wrong. We do not handle this problem within the basic link grammar framework; rather we treat these problems in a post-processing phase described in section 7.

### 3.7. Prepositions, adverbs, and complex noun phrases (`J`, `EV`, `EX`, `M`)

Prepositions have a `J+` connector; nouns have a `J-` connector.

> in on at for about: `J+`;
> dog: `{@A+} & Ds- & (Ss+ or SIs- or O- or J-)`;

These connections are used in prepositional phrases: ...*in the park*, ...*at the dog*. Prepositions also have `EV-` connectors; verbs have `EV+` connectors.

> hits: `Ss- & O+ & {@EV+}`;
> in at on: `J+ & EV-`;

The `EV+` on verbs has a multiple connection symbol, allowing it to take any number of `EV-` connections:

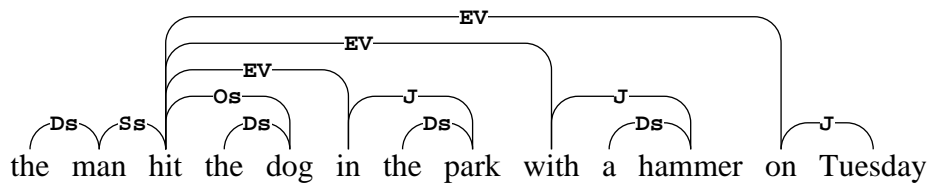

Prepositions also have `Mp-` connectors; nouns have `M+` connectors. The `M+` connectors on nouns also have a `@`, allowing nouns to take multiple prepositional phrases:

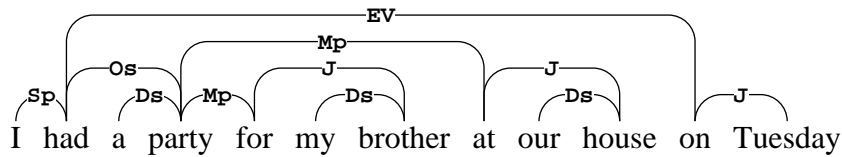

We should point out here that the linkage shown above is only one of a number that the program will find. In one linkage, *for*, *at* and *on* will be regarded as modifying *party*. In another case, *for* will connect to *party*, and *at* will connect to *brother*, and *on* will connect to *house*. Any of these prepositions may also connect to *had*. In long sentences, this can result in large numbers of linkages. To some extent, our program addresses this problem with heuristic rules for preferring some linkages over others, which we will not describe here. In many cases, though, the correct choice can only be made by a person, or program, possessing a great deal of semantic information: knowing, for example, that parties are often *on Tuesday*, while brothers are generally not. In short, choosing the correct linkage is largely a semantic matter; the job of a syntactic parser is to find all linkages which might possibly be correct.

We mentioned that nouns have `M+` connectors, used for prepositional phrases. These also allow nouns to take passive participles and present participles. Passive participles have `Mv-` connectors, present participles have `Mg-` connectors:

```
dog: {@A+} & Ds- & {@M-} & (Ss+ or SIs- or O- or J-);
chased: ((S- or T-) & O+) or V- or Mv-;
chasing: (GI- or Mg-) & O+;
```

This allows:

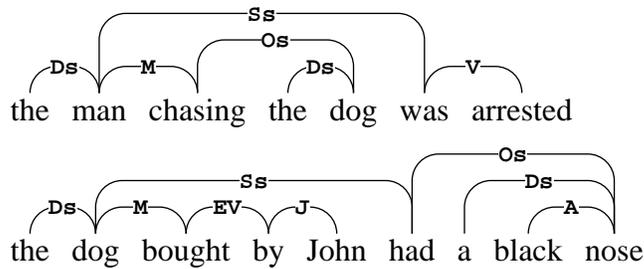

Recall that verbs have `EV+` connectors, used for prepositional phrases. These are also used for adverbs:

```
chased: (((S- or T-) & O+) or V- or Mv-) & {@EV+};
angrily quickly: EV-;
```

Adverbs may also be inserted before the verb; for this we give adverbs `EX-` connectors, and verbs `@EX-` connectors:

```
chased: {@EX-} & (((S- or T-) & O+) or V- or Mv-) & {@EV+};
angrily quickly: EV- or EX+;
```

Thus we allow

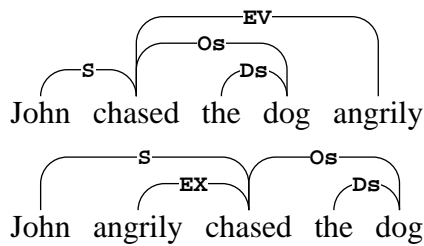

but do not accept

?John chased angrily the dog

although it is true that this construction is occasionally seen. Some adverbs may only be used before the verb, never after:

John apparently chases dogs
*John chases dogs apparently

To these adverbs we give `EX-` connectors but not `EV+` connectors. There are a number of other uses of adverbs handled by our on-line dictionary which we will not discuss here. Some examples are given in section A.

## 3.8. Conjunctions (`CL`, `CO`, `X`)

We mentioned above that complete clauses have `CL-` connectors. We will now show how this is achieved. Nouns (and pronouns) have `CL-` connectors &ed with the subject connectors:

> dog idea: {@A-} & Ds- & {@M+} & ((Ss+ & {CL-}) or SIs- or O- or J-);

A conjunction such as *because* has an `EV-` and a `CL+`. That is, it links to the verb of the clause before it (in the same manner as a prepositional phrase) and it links to the subject of the clause after it.

> because although while as: EV- & CL+;

A conjunction can also begin a sentence, followed by a subordinate clause, followed by a main clause. As before, the conjunction links to the `CL-` connector on the subject of the subordinate clause. It also links to a `CO-` connector on the subject of the main clause. The conjunction also has an `Xc+` connector; commas have an `Xc-` connector. (This is the only use of commas that will be explained here; our on-line dictionary includes a number of others.)

> because although while as: CL+ & (EV- or ({Xc+} & CO+));
> dog idea: {@A-} & Ds- & {@M+} & ((Ss+ & {{@CO-} & {CL-}}) or SIs- or O- or J-);
> ,: Xc-;

Note that the `CO-` on nouns has an `@`. This means it can take more than one *opener*:

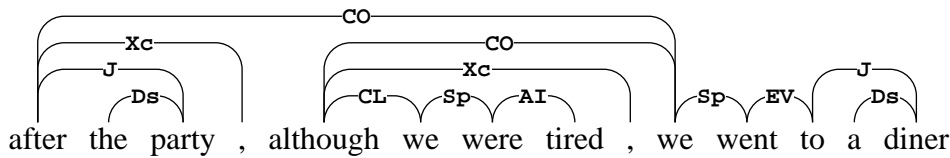

after the party , although we were tired , we went to a diner

The `CO-` on nouns is &ed with the `CL-`; this means that a subordinate clause can have an opener itself:

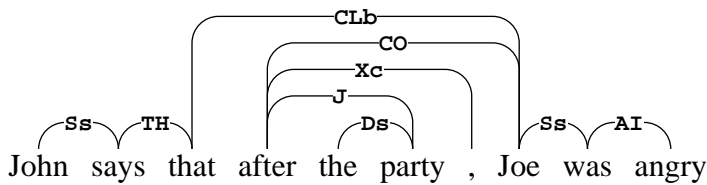

John says that after the party , Joe was angry

Some words, such as *before* and *after*, can act as conjunctions or prepositions:

> after before: (CL+ or J+) & (Mp- or EV- or ({Xc+} & CO+));

This, then, allows the following:

> We had a party after John passed his exam
> We had a party after the exam
> The party after the exam was a success
> ?The party after John passed his exams was a success
> After John passed him exams, we had a party
> After the exams, we had a party

As usual, our on-line dictionary handles more uses of conjunctions than are described here. Some of these uses are shown in section A.

### 3.9. Relative clauses (`B`, `C`, `Z`)

We have found several ways of dealing with the problem of relative clauses. The one we describe here may not be the simplest, but it works.

First let's consider sentences where a noun is the object of a relative clause:

> The dog John kicked died

To deal with this, nouns have a `B+` connector and transitive verbs have a `B-` connector.

> dog: `Ds- & {@M+ or B+} & ... ;`
> kicked: `(S- or T-) & (O+ or B-);`

Note that the `B+` on *dog* is optional. It is `&`ed with the `S+`, `SI-` and `O-` connectors, and also with the `D-` connector. This means that whether or not a noun takes a relative clause, its other requirements are unaffected: it still needs a determiner, and it still needs to be an ordinary subject or an object. Note also that the `B+` on nouns is `or`ed with the `@M+`, which, you may recall, is used for past and present participles and for prepositional phrases. In other words, in our system, a noun can not take both a relative clause and a prepositional phrase. The following sentences are thus rejected:

> ?The dog chasing me John bought died
> ?The dog Joe kicked bought by John died

Note also the ordering in the previous expression for *dog*. The `B+` is before the `Ss+` (its ordering relative to `-` connectors is irrelevant). This means that the program will accept

> The dog John kicked died

but will reject

> *(?) The dog died John kicked

(This suffices in the vast majority of cases, but occasionally one does find a sentence like *the dog died which John kicked*!)

The `B-` on *kicked* is `or`ed with its `O+` connector; that means that if *kicked* is connecting back to a noun in a relative clause, it cannot also be connecting forwards to another object:

> *The dog John kicked the cat died

In object-type relative clauses, *who* (or *that*, or *which*) is optional. This we allow by giving nouns a `C+` connector, optionally `or`ed with the `B+` connector. *Who* is given a `C-` connector.

> dog: `{@A+} & Ds- & {@M+ or ({C+} & B+)} & ... ;`
> who: `C-;`

This allows both of the following:

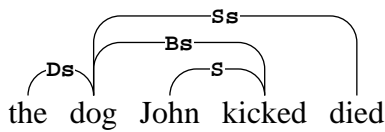

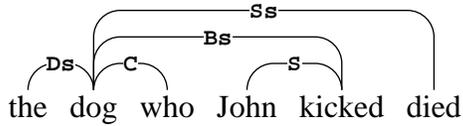
the dog who John kicked died

How about subject-type relative clauses, such as

The dog who chased cats died

We use B connections for this as well. Nouns already have B+ connections, as described above. On verbs, we now install B- connections ored with the S- connection. The standard singular/plural subscripting system is used here as well.

chases: {@EX-} & (Ss- or Bs-) & (B- or O+) & {@EV+};
chase: {@EX-} & (Sp- or Bp- or I-) & (B- or O+) & {@EV+};

Thus we can correctly judge the following:

The dog who chases cats died
*The dogs who chases cats died

The B+ on nouns is &ed with the C+, so *who* can be used just as it is with object-type relative clauses. But now, there's a complication! With subject-type the *who* is not optional, as with object-type clauses; rather it is obligatory. We have to find a way of preventing

*The dog chased cats died

Therefore, we introduce the Z connector.

chased: (Ss- or (Z- & Bs-)) & (O+ or B-);
who: C- & {Z+};

Now, a sentence like

*The dog chased cats died

will not be permitted. The B+ on *dog* will try to connect to the Bs- on *chased*, but it will not be allowed to, because *chased* must also make a Z connection in order for its B+ to be usable. With a sentence like

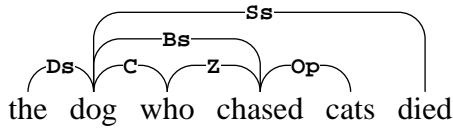
the dog who chased cats died

the *who* provides the Z connection; the C- on *who* connects to the C+ on *dog*, and everybody's happy.

*Which* and *that* have the same connector expressions as *who* (at least, they are the same in this respect). This means that the program does not distinguish between person-nouns, which should take *who*, and thing-nouns, which should take *which*. It could quite easily do so, with the use of subscripts, but we feel that this is a semantic matter rather than a syntactic one.

Certain expressions can be added to, or substituted for, *who* in both object-type and subject-type relative clauses:

The man I think John hit died
The man I think hit John died
The man who I think John hit died
The man who I think hit John died

We can allow this simply by giving such verbs `Z+` connectors:

    think: (S- or I-) & (O+ or B- or TH+ or CL+ or Z+);

This then allows

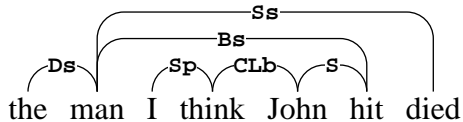

### 3.10. Questions (`Q`)

There are many different kinds of questions. In *who-subject* questions, *who* simply acts like the subject of a sentence:

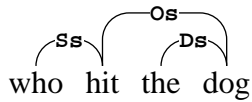

This is easily dealt with: we simply give *who* a `S+` connector. This is `or`ed with the relative clause connectors discussed above.

    who: (C- & {Z+}) or S+;

In *who-object* type questions, the subject and verb are inverted. To accomplish this, we use the `SI` connector described earlier, in conjunction with a new connector called `Q`:

    who: (C- & {Z+}) or S+ or (Q+ & B+);
    did: (S- or (SI+ & {Q-})) & I+;

This allows

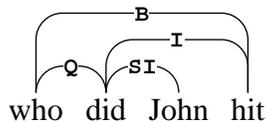

Note that if the `SI+` connector on *did* is used, then it may also form a `Q` connection backwards; this is optional. However, if a `Q` connection is formed, then the `SI` connector on *did must* be used. Looking at the expression for *who*, we see that the `Q` connector is `&`ed with an `B+` connector, which was also used with relative clauses. All this ensures the following: 1) If a sentence begins with *who*, then *who* must either form an `S` connection with a verb or a `Q` connection with an Invertible verb. 2) If it forms a `Q` connection, there must be subject-verb inversion, and there must be a transitive verb in the sentence to satisfy the `B+` on *who*. 3) In a sentence with subject-verb inversion, there need not be a *who*, but if there is a transitive verb in the sentence with no object following it, then there must be a *who* at the beginning. Therefore the following sentences will be correctly judged.

>     Who hit John?
>     *Who John hit?
>     *Who John hit the cat?
>     *Who did John hit the cat?
>     Who did John hit?
>     *Did John hit?
>     Did John hit the cat?

There is a serious problem with this system. The following completely bogus construction is accepted:

>     *Who did John die Joe hit

We solve this problem in the post-processing phase.

As mentioned above, some verbs take question words; an R connector is used for this. However, in such a case, there can be no subject-verb inversion.

>     I wonder who John hit.
>     I wonder who hit John.
>     *I wonder who did John hit.

Therefore *who* is given the following expression:

>     who: (C- & {Z+}) or ({R-} & S+) or ((R- or Q+) & B+);

This allows

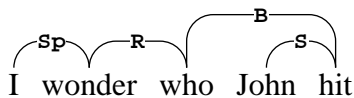
I wonder who John hit

Remember that it is the Q+ connector on *who* which triggers subject-verb inversion. Therefore, this says that if an R connection is made, a Q connection can not be made; therefore subject-verb inversion can not take place.

With no further tinkering necessary, the system also correctly judges the following kinds of complex sentences:

>     Who do you think John hit?
>     Who do you think hit John?
>     I wonder who you think John hit.
>     I wonder who you think hit John.
>     *I wonder who do you think hit John.
>     *I wonder you think John hit.
>     *Who you think hit John?

Each question word has slightly different properties. *What* is similar to *who*, but it can not be used in relative clauses:

>     *The dog what John bought died

Unlike *who*, *what* and *which* can act as determiners, while also acting as question words:

What movie did you see?

The problem here is that in such a sentence, *what* fulfills both the obligatory demands of nouns; in the sentence above, *movie* requires neither a determiner or a main subject-object function. (The `B-` connector on *see* is taken by *what*.) Therefore, we give nouns `U-` connectors, which are ored with everything else on the noun except `A-` (adjective) connectors. We give *what* and *which* `U+` connectors.

    dog: {@A-} & ((Ds- & {@M+ or ({C+} & Bs+)} &
          ((Ss+ & {{@CO-} & {CL-}}) or SIs- or O- or J-)) or U-);
    what: {U+} & (((R- or Q+) & B+) or ({R-} & S+));

*When*, *where*, and *why* can start questions, but unlike *who*, they do not take objects. They have `Q+` connectors, like *who*, forcing subject-verb inversion; but they do not have `B+` connectors. For indirect questions (*I wonder why John left*) `CL` connections are used.

    when where why how: Q+ or (R- & CL+);

*When* can act as a conjunction; *why*, *where* and *how* can not:

    when: Q+ or (CL+ & (R- or EV-));
    where why how: Q+ or (R- & CL+);

As with adverbs and conjunctions, there are a number of subtleties in the use of question words which the program handles well, but which we will not discuss here nor include in the abridged dictionary. Some examples are given in section A.

### 3.11. The wall

There is a problem with the method of handling question inversion described above: nothing prevents question inversion from taking place in relative clauses. This means that the following are all accepted:

> Did John hit the dog
> *The dog did John hit died
> When did John hit the dog
> *The dog when did John hit died

To solve this, we invented something called *the wall*. The wall can best be thought of as an invisible word which the program inserts at the beginning of every sentence, and which has a connector expression like any other word. The wall has only optional connectors; it does not have to connect to anything in the sentence. (It is therefore exempt from the 'connectivity' rule which states that in a valid sentence all words must be connected[5].) However, there are words in the dictionary whose demands can be satisfied by connectors on the wall. We use this to ensure that question inversion only occurs at the beginning of sentences. So, the wall has a `Q+` connector. Invertible verbs such as *did* and *has* have a `Q-` connector, as mentioned above. However, instead of making this connector optional – as we originally described it – we can now make it obligatory:

---

[5]Actually, the wall is implemented by modifying the disjuncts on the words in the sentence and then applying the standard algorithm for finding a linkage. A new disjunct of the form `(()(WA))` is added to the wall. Every disjunct on the first word of the sentence is duplicated, and in the duplicate of each one a `WA-` connector is added.

```
    did: ((Q- & SI+) or S-) & I+;
    has: ((Q- & SIs+) or Ss-) & T+;
```

This means that when question inversion takes place, a `Q` connection must be made. It can either be made to a question word, such as *when*:

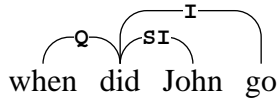

or it can be made to the wall (denoted "/////"):

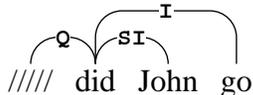

This prevents the first invalid sentence given above, *the dog did John hit died*. How about the second, *The dog when did John hit died*? To prevent this, we give the wall a `W+` connector, and we give question words a `W-` connector anded with their `Q` connectors. This means that when question words are used in questions, they must be able to connect to the wall:

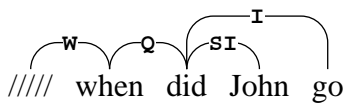

We also use this same system for imperatives. Every infinitive verb, as well as having an `I`-connector, also has a `W-`. This means any infinitive verb can begin a sentence, without a subject:

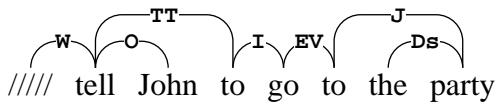

### 3.12. Some problems with questions

There are several problems with our handling of questions. One has been mentioned already the incorrect construction

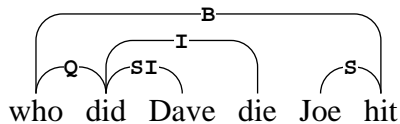

which is allowed by the dictionary included here. Other problems relate to constructions such as:

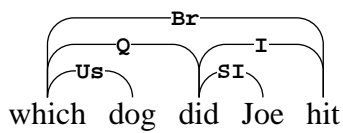

In this sentence, there is no connection between *dog* and *hit*. This in itself is unsatisfactory, since we try to make the grammar form links between words which are logically related, and a transitive verb is surely logically related to its object. This especially becomes a problem in sentences like *Which man does John think likes Mary*, where subject-verb agreement is necessary between *man* and *likes*.

Since there is no direct link between these two words, the usual means of ensuring subject-verb agreement – with subscripts – is not possible. A similar problem arises in constructions such as *How certain is John that Mary is coming?*.

To solve these problems, we devised another approach to the problem of questions, which we use in the on-line dictionary. This solution makes considerable use of post-processing. We will not describe it here; however, we offer a general discussion of post-processing mechanisms in section 7. (This is virtually the only case where the abridged dictionary and the on-line dictionary contain different solutions to the same problem. In almost all other respects, the definitions in the on-line dictionary can be regarded as extensions and refinements of those included here, rather than alternatives.)

## 4. The algorithm

Our algorithm for parsing link grammars is based on dynamic programming. Perhaps its closest relative in the standard literature is the dynamic programming algorithm for finding an optimal triangulation of a convex polygon [3, p. 320]. It tries to build up a linkage (which we'll call a *solution* in this section) in a top down fashion: It will never add a link (to a partial solution) that is above a link already in the partial solution.

The algorithm is most easily explained by specifying a data structure for representing disjuncts. A disjunct $d$ has pointers to two linked lists of connectors. These pointers are denoted $left[d]$ and $right[d]$. If $c$ is a connector, then $next[c]$ will denote the next connector after $c$ in its list. The next field of the last pointer of a list has the value NIL.

For example, suppose the disjunct $d =$ ((D,O) ()) (using the notation of section 2). Then $left[d]$ would point to the connector O, $next[left[d]]$ would point to the connector D, and $next[next[left[d]]]$ would be NIL. Similarly, $right[d] =$ NIL.

To give some intuition of how the algorithm works, consider the situation after a link has been proposed between a connector $l'$ on word $L$ and a connector $r'$ on word $R$. (The words of the sequence to be parsed are numbered from 0 to $N - 1$.) For convenience, we define $l$ and $r$ to be $next[l']$ and $next[r']$ respectively. The situation is shown in the following diagram:

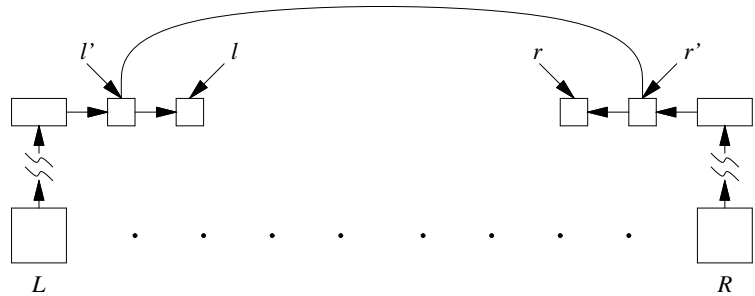

How do we go about extending the partial solution into the region strictly between $L$ and $R$? (This region will be denoted $(L, \ldots, R)$.) First of all, if there are no words in this region (*i.e.* $L = R + 1$) then the partial solution we've built is certainly invalid if either $l \neq$ NIL or $r \neq$ NIL. If $l = r =$ NIL then this region is ok, and we may proceed to construct the rest of the solution.

Now suppose that the region between $L$ and $R$ contains at least one word. In order to attach the words of this region to the rest of the sentence there must be at least one link either from $L$ to some word in this region, or from $R$ to some word in this region (since no word in this region can link to a word outside of the $[L, \ldots, R]$ range, and something must connect these words to the rest of the sentence).

Since the connector $l'$ has already been used in the solution being constructed, this solution must use the rest of the connectors of the disjunct in which $l'$ resides. The same holds for $r'$. The only connectors of these disjuncts that can be involved in the $(L, \ldots, R)$ region are those in the lists beginning with $l$ and $r$. (The use of any other connector on these disjuncts in this region would violate the ordering requirement.) In fact, all of the connectors of these lists must be used in this region in order to have a satisfactory solution.

Suppose, for the moment, that $l$ is not NIL. We know that this connector must link to some disjunct on some word in the region $(L, \ldots, R)$. (It can't link to $R$ because of the exclusion rule.) The algorithm tries all possible such words and disjuncts. Suppose it finds a word $W$ and a disjunct $d$ on $W$ such that the connector $l$ matches $left[d]$. We can now add this link to our partial solution. The situation is shown in the following diagram.

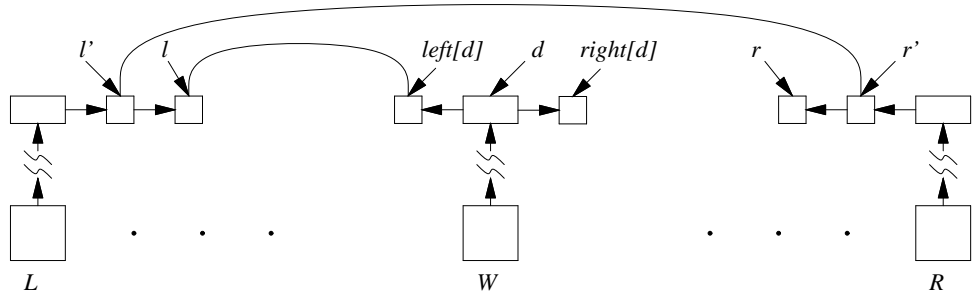

How do we determine if this partial solution can be extended to a full solution? We do this by solving two problems similar to the problem we started with. In particular, we ask if the solution can be extended to the word range $(L, \ldots, W)$ using the connector lists beginning with $next[l]$ and $next[left[d]]$. We also ask if the solution can be extended to the word range $(W, \ldots, R)$ using the connector lists beginning with $right[d]$ and $r$. Notice that in the latter case, the problem we are solving seems superficially different: the boundary words have not already been connected together by a link. This difference is actually of no consequence because the pair of links ($L$ to $R$ and $L$ to $W$) play the role that a direct link from $W$ to $R$ would play: (1) they separate the region $(W, \ldots, R)$ from all the other words, and (2) they serve to connect the words $W$ and $R$ together.

We need to consider one other possibility. That is that there might be a solution with a link between words $L$ and $W$ *and* a link between words $W$ and $R$. (This results in a solution where the word/link graph is cyclic.) The algorithm handles this possibility by also attempting to form a link between $right[d]$ and $r$. If these two match, it does a third recursive call, solving a third problem analogous to our original problem. In this problem the word range is $(W, \ldots, R)$ and the connector lists to be satisfied begin with $next[right[d]]$ and $left[r]$.

A very similar analysis suffices to handle the case when $l$ is NIL.

Having explained the intuition behind our algorithm, we can now describe it more formally. In the pseudocode[6] below, the boolean function MATCH takes two connectors and returns *true* if the names of these connectors match, and *false* otherwise.

---

[6]In this notation, the grouping of statements is indicated by the level of indentation. More details can be found in Cormen *et al.*[3, pp. 4-5].

PARSE
1  $t \leftarrow 0$
2  **for** each disjunct $d$ of word 0
3      **do if** $left[d] = $ NIL
4          **then** $t \leftarrow t + $ COUNT$(0, N, right[d], $ NIL$)$
5  **return** $t$

COUNT$(L, R, l, r)$
1  **if** $L = R + 1$
2    **then if** $l = $ NIL and $r = $ NIL
3        **then return** 1
4        **else return** 0
5    **else** $total \leftarrow 0$
6        **for** $W \leftarrow L + 1$ **to** $R - 1$
7          **do for** each disjunct $d$ of word $W$
8              **do if** $l \neq $ NIL and $left[d] \neq $ NIL and MATCH$(l, left[d])$
9                  **then** $leftcount \leftarrow $ COUNT$(L, W, next[l], next[left[d]])$
10                **else** $leftcount \leftarrow 0$
11              **if** $right[d] \neq $ NIL and $r \neq $ NIL and MATCH$(right[d], r)$
12                **then** $rightcount \leftarrow $ COUNT$(W, R, next[right[d]], next[r])$
13                **else** $rightcount \leftarrow 0$
14              $total \leftarrow total + leftcount * rightcount$
15              **if** $leftcount > 0$
16                **then** $total \leftarrow total + leftcount * $ COUNT$(W, R, right[d], r)$
17              **if** $rightcount > 0$ and $l = $ NIL
18                **then** $total \leftarrow total + rightcount * $ COUNT$(L, W, l, left[d])$
19          **return** $total$

The function COUNT takes as input indices of two words ($L$ and $R$), and a pair of pointers to lists of connectors ($l$ and $r$). It is easy to verify that these parameters satisfy the following conditions: $0 \leq L < R \leq N$, and $l$ points to a connector on the right list of a disjunct on word $L$, and $r$ points to a connector on the left list of a disjunct on word $R$. COUNT returns a number. This is the number of different ways to draw links among the connectors on the words strictly between $L$ and $R$, and among the connectors in the lists pointed to by $l$ and $r$ such that these links satisfy the following conditions:

1. The are planar (they do not cross), and they also satisfy the ordering and exclusion meta-rules.

2. There is no link from a connector of the $l$ list to one of the $r$ list.

3. The requirements of each word strictly between $L$ and $R$ are satisfied by the chosen links.

4. All of the connectors in the lists pointed to by $l$ and $r$ are satisfied.

5. The links either connect all the words $[L,\ldots,R]$ together, or they form two connected components: $[L,\ldots,M]$ and $[M+1,\ldots,R]$ (for some $M$ satisfying $L \leq M < R$).

We will leave the induction proof that the algorithm is correct (that these five conditions always hold) as an exercise for the reader.

The pseudocode follows the basic outline of the intuitive discussion, but is slightly more sophisticated[7]. Lines 8 through 18 are done for every choice of intermediate word $W$ and every disjunct $d$ on that word. Lines 8 through 10 assign to *leftcount* the number of ways to complete the solution using a link from $L$ to $W$. Lines 11 through 13 assign to *rightcount* the number of ways to complete the solution using a link from $W$ to $R$. The product of these two (computed on line 14) is the number of ways of using both of these links. Lines 15 through 18 handle the cases where only one of these pairs of words are linked together.

The entire function is symmetrical with respect to left-right reversal except for the "$l =$ NIL" test on line 17. This is necessary to ensure that each solution is generated exactly once. For example, if this were deleted than the following solution would be generated twice:

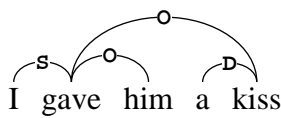
I gave him a kiss

In trying to link together the range [gave,...,kiss] it would generate this solution once when the partitioning word ($W$) is *him* and once when the partitioning word is *a*. The problem is that the same set of links is being generated in two different orders. If there was a canonical ordering imposed on the links, the problem would be solved. If $l \neq$ NIL then we know that (if there is any solution) then the connector $l$ must be used eventually. The test insures that this link is generated first. Therefore the above solution would be generated only when the partitioning word is *him*.

Once one believes that COUNT satisfies the invariants above, it is easy to check that PARSE returns the number of ways to parse the given sequence of words. Consider a top level call to COUNT. Condition 5 implies that the chosen links must connect all the words together. (Since $r$ is the empty list, there must be two components $[0,\ldots,N-1]$ and $[N]$.) Conditions 1 and 2 guarantee planarity, exclusion and ordering. Finally, condition 3 guarantees satisfaction. Each iteration of the main loop of PARSE tries to use a different disjunct on the first word of the sentence. Since exactly one of these must be used in any valid arrangement of links, this algorithm enumerates all possibilities.

For simplicity we have ignored multi-connectors in our description of the algorithm. These can easily be incorporated into this framework. If one end of a link is formed with a multi-connector we need to make two recursive calls to the region spanned by that link. One of the calls (as above) advances the pointer (pointing to the multi-connector). An additional call is made that does not advance this pointer. This allows the multi-connector to be used again in links below the current link. If both ends of the link are multi-connectors, then three recursive calls are required.

---

[7]The intuitive discussion describes the process of building up a single solution. The algorithm does not remember any links at all. A separate process, not described in this paper, is used to construct the links. If you halt the program at some point and look at what been considered, it isn't just a single partial solution. It's a set of partial solutions. This is because there may be many ways of satisfying a region with particular boundary conditions. The algorithm considers all such ways.

The algorithm described has an exponential worst-case running time as a function of $N$. This can easily be transformed into an efficient dynamic programming algorithm by using *memoization* ([3, p. 312]). To install this, the function COUNT is modified in two ways. It is made to store into a hash table its answer (right before it returns). It is also modified so that the first thing it does is to look in the hash table to see if a prior call has already computed the desired value.

The running time is now bounded by the number of different possible recursive calls multiplied by the time used by each call. A recursive call is completely determined by specifying the pointers $l$ and $r$. (These uniquely determine $L$ and $R$.) The cost of a given call is bounded by the total number of disjuncts in the sequence of words.

If we let $d$ be the number of disjuncts and $c$ be the number of connectors, then the running time is $O(c^2 d)$. For a fixed link grammar, $d = O(N)$ and $c = O(N)$, so the running time is $O(N^3)$.

## 5. Speeding it up

When we implemented and tested our link parsing algorithm on a non-trivial dictionary, we were disappointed by its performance. It turned out that in a dictionary with significant coverage of English grammar the number of disjuncts on many words gets rather large [8].

After studying what was happening, it became apparent that other techniques might be used (in conjunction with our algorithm) to improve the performance of the system. We devised and implemented several schemes: some proved to be extremely effective while others were useless. In this section we describe this work. We will start with the methods that were found to be effective, and proceed to the ones that were not.

### 5.1. Pruning

Our first approach is based on the following observation: In any particular sequence of words to be parsed, most of the disjuncts are irrelevant for the simple reason that they contain a connector that does not match any other connector on a word in the sequence. To be more precise, suppose that a word $W$ has a disjunct $d$ with a connector C in its right list. If no word to the right of $W$ has a connector (pointing to the left) that matches C, then the disjunct $d$ cannot be in any linkage. This disjunct can therefore be deleted without changing the set of linkages. Deleting such a disjunct is called a *pruning step*. *Pruning* consists of repeating the pruning step until it can no longer be applied.

The set of disjuncts left (after pruning is complete) is independent of the order in which the steps are applied. (The pruning operation has the Church-Rosser property.) We therefore choose an ordering that can be efficiently implemented. It would be ideal if we could achieve a running time for pruning that is linear in the number of connectors. The scheme we propose satisfies no useful a-priori bound on its running time, but in practice it appears to run in linear time.

A series of sequential passes through the words is made, alternating between a left-to-right pass and a right-to-left pass. The two types of passes are analogous, so it suffices to describe the left-to-right pass. The pass processes the words sequentially, starting with word 1. Consider the situation after words $1, \ldots, W - 1$ have been processed. A set $S$ of connectors has been computed. This is the set of connectors that exists on the right lists of the disjuncts of words $1, \ldots, W - 1$ that have not been deleted. To process word $W$, we consider each disjunct $d$ of $W$ in turn. For each connector $c$ on the left list of $d$, we search the set $S$ to see if it contains a connector that matches $c$. If one of the connectors of $d$ matches nothing in $S$, then we apply the pruning step to $d$ (we remove $d$). Each right connector of each remaining disjunct of $W$ is now incorporated into the set $S$. This completes the processing of word $W$.

The function computed by this left-to-right pass is idempotent, which is a high-flautin' way of saying that if you've done it, then immediately doing it again will have no effect. Therefore if (as we alternate left-to-right and right-to-left passes) a pass (after the first one) does nothing, then all further passes will do nothing. This is how the algorithm decides when to stop.

---

[8] For example, in the abridged dictionary singular nouns have 66 disjuncts, and plural nouns have 130. In our on-line dictionary (of September 3, 1991) these numbers are 213 and 391 respectively. Some words have many more. For example *time* as a noun has 770 disjuncts.

It is also clear that if a pruning step is possible, then it will be discovered either in a left-to-right pass or a right-to-left pass. Thus, what we've described is a correct implementation of pruning.

The data structure used for the set $S$ is simply a hash table, where the hash function only uses the initial upper-case letters of the connector name. This ensures that if two connectors get hashed to different locations, then they definitely don't match.

Although we know of no non-trivial bound on the number of passes, we have never seen a case requiring more than five. Below is a typical example. The sentence being parsed is: *Now this vision is secular, but deteriorating economies will favor Islamic radicalism.* This is the last sentence of the second transcript in appendix B. The table below shows the number of disjuncts (as it evolves) for each of the words of this sentence. (The first number is for the wall. Of course the comma also counts as a word.) A fourth pass is also made, which deletes no disjuncts.

|  |  |  |  |  |  |  |  |  |  |  |  |  |  |
|---|---|---|---|---|---|---|---|---|---|---|---|---|---|
| Initial: | 3 | 12 | 18 | 391 | 296 | 9 | 10 | 3 | 81 | 391 | 20 | 423 | 104 | 391 |
| after L→R | 3 | 8 | 11 | 67 | 160 | 6 | 10 | 3 | 81 | 163 | 8 | 381 | 25 | 357 |
| after R→L | 3 | 6 | 5 | 25 | 21 | 3 | 3 | 3 | 25 | 25 | 3 | 25 | 4 | 12 |
| after L→R | 3 | 6 | 5 | 25 | 21 | 3 | 3 | 3 | 25 | 21 | 3 | 21 | 3 | 8 |

## 5.2. The fast-match data structure

Looking at the pseudo-code for the algorithm in section 4 reveals the following fact: In a particular call to COUNT the only disjuncts of a word $W$ that need to be considered are those whose first left connector matches $l$, or whose first right connector matches $r$. If there were a fast way to find all such disjuncts, significant savings might be achieved. The fast-match data structure does precisely this.

Before beginning the search we build two hash tables for each word of the sequence. These are called the *left* and *right* tables of the word. These hash tables consist of an array of buckets, each of which points to a linked list of pointers to disjuncts. Each disjunct with a non-null right list is put into the right hash table, and each disjunct with a non-null left list is put into the left hash table (many disjuncts will be stored in both tables). The hash function for a right table depends only on leading upper-case letters of the name of the first connector of the right list of the disjunct being hashed. (The hash function for the left table is analogous.)

These tables allow us to quickly solve the following problem: given a connector $l$, and a word $W$, form a set of disjuncts of $W$ [the first connector of the left list of which] might possibly link to $l$. (Of course, this is only going to be useful if the set formed is much smaller than the total number of disjuncts. This is usually the case, since the connectors of a word are distributed roughly evenly between many different types.)

To apply this data structure, the algorithm is modified as follows: When trying an intermediate word $W$, compute the two sets of disjuncts mentioned in the previous paragraph. Compute the union of these two sets and (while working on word $W$) restrict attention to these disjuncts.

Assuming all types of connectors are used with equal frequency, this technique speeds up the algorithm by a factor equal to the number connector types. Experiments show that the method is very effective.

Later we will show how the fast-match data structure can be augmented to elegantly take advantage of information computed by power pruning.

### 5.3. Power pruning

Power pruning is a refinement of pruning that takes advantage of the ordering requirement of the connectors of a disjunct, the exclusion rule, and other properties of any valid linkage.

Say that a connector is *shallow* if it is the first connector in its list of connectors. A connector that is not shallow is *deep*. Our analysis begins with the following observation:

O: In any linkage, successive connectors on a connector list of a disjunct of word $w$ must connect to words that are monotonically closer to $w$.

To take advantage of this, we endow each connector with an additional (dynamically changing) piece of data called the *nearest-word*. Its meaning is the following: There is no linkage in which a connector links to a word that is closer than its nearest-word. These fields are initialized to the nearest word allowed by observation O. (For example, a shallow connector on a right list of three connectors starts with its nearest-word three to the right of the current word.) As power pruning proceeds, the nearest-words increase in distance (from the word containing the connector). When a nearest-word goes outside the sequence of words the corresponding disjunct may be deleted.

Here are several local (easy to test) criteria that can be used to prove that a proposed link cannot be in any linkage:

1. The nearest-word criterion must be satisfied for both connectors of a link.

2. There can be no link between two deep connectors. (Proof: If two deep connectors are linked, then there is no place for the corresponding shallow connectors to link.)

3. The two connectors of a link between two neighboring words must be the last connectors of their respective lists. (Proof: any further connectors on either of these two lists would have nowhere to link.)

4. The two connectors of a link between two non-neighboring words cannot both be the last connectors of their respective lists (unless one of them is a multi-connector). (Proof: How are the words between these two non-neighboring words going to be connected to the rest of the sentence?)

We can now describe the power pruning algorithm. Just as with pruning, we make an alternating sequence of left-to-right and right-to-left passes. Consider processing the word $W$ in a left-to-right pass. We keep a collection of sets, one for each word to the left of $W$. Each set is represented as a hash table that stores the connectors that point to the right from each word. For each disjunct of $W$ we recompute the nearest-word fields of its left pointing connectors, according to the four local conditions above, and observation O. This is done by starting from the deepest left connector of the disjunct, searching from its nearest-word to the left, looking for a connector to which it might match. Once this match is found, the nearest-word field is updated, and the preceding connector on the list is similarly updated. This process is applied to all the connectors of the disjunct. If one of the nearest-word fields walks past the first word of the sequence, the disjunct may be eliminated. After processing each disjunct of word $W$, the hash table is built for the remaining right connectors of $W$, completing the processing of $W$.

This method is idempotent – repeating a left-to-right pass immediately would do nothing. Therefore, just as in pruning, the process terminates when a pass (after the first one) changes no nearest-word fields and deletes no disjuncts.

The following diagram shows what happens to the example above is after power pruning is applied.

| After pruning: | 3 | 6 | 5 | 25 | 21 | 3 | 3 | 3 | 25 | 21 | 3 | 21 | 3 | 8 |
| --- | --- | --- | --- | --- | --- | --- | --- | --- | --- | --- | --- | --- | --- | --- |
| After P.P.: | 2 | 6 | 1 | 13 | 2 | 2 | 2 | 2 | 2 | 6 | 1 | 4 | 2 | 1 |

There is a beautiful way to incorporate the nearest-word information into the fast-match data structure. In each bucket of every fast-match hash table, the disjuncts are sorted by increasing nearest-word fields of their first connector. (The disjunct whose connector can link the nearest is the first, etc.) Now, when forming the list of connectors of a word $W$ that might match to a connector $r$ ($l$), we also can make use of the word $R$ ($L$) on which $r$ ($l$) resides. The list formed is not simply a copy of the hash bucket, but rather it is a portion of this list – the portion containing connectors that do not violate the nearest-word constraint. Since the list of disjuncts in a bucket in the hash table are sorted by nearest-word value, the desired portion of the list is an initial portion of it, and it can easily be constructed. Other than this difference, the fast-match data structure operates as described earlier.

The effect of this technique is to dynamically change the set of disjuncts that "exist" for the purpose of attempting to build a linkage. This implicit dynamic pruning of the disjuncts takes place with no extra cost in time or space.

The nearest-word information can also be used to directly restrict the range of intermediate words. We only need to consider intermediate words $W$ in the range from the nearest-word of $l$ to the nearest-word of $r$. If this range is empty, then there is no valid linkage to be found in this branch of the search.

### 5.4. Clever ideas that don't seem to work

In the current implementation pruning and power pruning take roughly as much time as the exhaustive search for a linkage. (See appendix C.) This renders irrelevant any further improvements in the searching algorithm. However, even after the development of pruning, the searching step was excruciatingly show on long sentences. Before stumbling on the mighty fast-match and power pruning combination, several other ineffective techniques were tried.

**Folding**  This idea is similar to that used by Appel and Jacobson [2] to compress a scrabble dictionary. In our application, however, the goal is to save time rather than space. Suppose that, just for the sake of illustration, all of the right connector lists of the disjuncts of a word ended with the same sublist of two connectors. We could replace the multiple copies of this sublist by a single one. Instead of having a collection of separate lists, we have a tree. This saves some space, but the main advantage here is obtained when using this modified data structure in conjunction with the search algorithm of the previous section. Because of memoization, the coalescing of the lists means that many recursive calls that were were considered different by the old algorithm are now recognized for what the are – the same problem. The method turned out – rather mysteriously – to have almost no effect on the running time.

**Recursive Pruning**  The stunning success of pruning implies that a good way to search for a linkage is to prune before you search. Since searching for a linkage in a long sentence is reduced by the recursive algorithm to finding linkages (or proving there are none) in shorter sentences, it is natural to ask "why not prune at the beginning of each recursive call". Indeed, the idea makes sense, and we tried it. Unfortunately, it had almost no effect on the running time. This is perhaps because (1) the first pruning step does much more than subsequent ones, and (2) pruning takes a lot of time, and cannot be done indiscriminately.

**Alternative search orders**  Another standard technique to speed up exhaustive search is to adjust the search order. In our case, a solution consists of a valid linkage in the region between $L$ and $W$ and also in the region between $W$ and $R$ (harking back to the notation of section 4). If one of these fails, then the whole thing fails. Isn't it natural, therefore, to think that searching the smallest range first would be better than searching the left one first (like the algorithm above)? This reasoning is correct, assuming that the small ones and the large ones fail equally often, and that the small ones take less time than the large ones. You see, if the small one does fail, then you've avoided doing a (probably more expensive) search of the large one. It may be natural, but it's wrong. Searching the smallest region first ended up slowing the program down by about a factor of two.

## 6. Coordination phenomena

Various coordination phenomena do not fit naturally into the framework of link grammars. We have devised a method for automatically transforming the given link grammar into another one that captures the desired phenomena. Although not perfect, the ideas described here have proven to be very effective.

Our discussion will focus on the word *and*, although the ideas apply to the use of *or*, *either*···*or* *neither*···*nor*, *both*···*and*···, *not only*···*but*···[9]. First, we propose a definition for the grammatical uses of *and* in the context of link grammars. Then we describe our system for generating extra disjuncts on words that accommodates the vast majority (in real texts) of these uses.

### 6.1. What is a valid way of using *and*?

We are not aware of any elegant definition of the grammatical uses of *and* in any linguistic system. We begin by proposing a simplistic definition within the framework of link grammars. Then we'll mention a few problems with the definition, and suggest an improvement. The second definition is the one used in our system. It has drawbacks, but on balance it has proven to be remarkably effective.

Given a sequence $S$ of words containing the word *and*, a *well-formed and list* $L$ is a contiguous subsequence of words satisfying these conditions:

- ¶ There exists a way to divide $L$ into components (called *elements* of the well-formed *and* list) such that each element is separated from its neighboring elements by either a comma or the word *and* (or both). (The comma and the *and* are not part of the element.) The last pair of elements must be separated by *and* (or a comma followed by *and*). For example, in *the dog, cat, and mouse ran*, *dog*, *cat*, and *mouse* are the elements of the well-formed *and* list *dog, cat, and mouse*.

- ¶ Each of the sequences of words obtained by replacing $L$ (in $S$) by one of the elements of $L$ is a sentence of the link grammar.

- ¶ There is a way of choosing a linkage for each of these sentences such that the set of links outside of the *and* list elements are exactly the in all of the sentences, and the connectors joining the sentence with its *and* list element are the same. In other words, if we cut the links connecting the element to the rest of the sentence, remove that element from the sentence, and replace it by one of the other elements, then the cut links can be connected to the element so as to create a valid linkage.

The sequence $S$ is grammatical if each instance of *and* in it is part of a well-formed and list.

---

[9]Our current implementation handles all but the last two of these.

For example, consider the sentence *We ate popcorn and watched movies on TV for three days.* The the phrase *ate popcorn and watched movies on TV* is a well-formed and list because it can be divided into elements *ate popcorn* and *watched movies on TV*, which satisfy all of the conditions above. The following two linkages show this. Note that in both cases the element is attached to the rest of the sentence with an S to the left and a EV to the right.

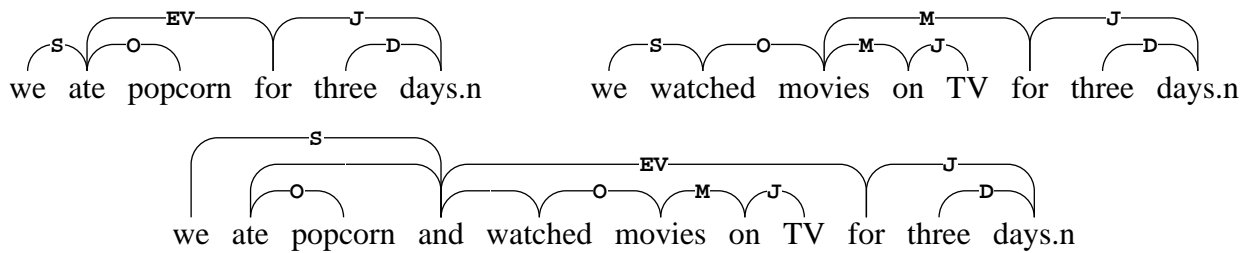

There is a major problem with this definition. It contains no requirement that the words of an element of an *and* list be connected to each other, nor be related in any way (other than being contiguous). This allows many clearly ungrammatical sentences to be accepted, and generates numerous spurious linkages of correct sentences. For example, the following two linkages:

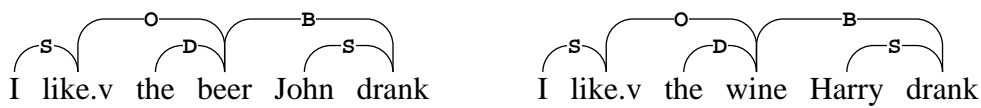

imply (according to the above definition) that

>    *I like the beer John and wine Harry drank is a valid sentence.

We have two techniques to limit the set of sentences deemed grammatical by this rule. The first is to simply restrict the types of connectors that can connect the element of the *and* list to the rest of the sentence[10]. The second method is to restrict the definition of a well-formed *and* list. Say that a well-formed *and* list is a *strict and list* if it also satisfies the following condition.

¶  Each element must be connected to the rest of the sentence through exactly one of its words. (It may use many connectors.)

Although the judgements rendered by this solution are almost always correct, it is sometimes wrong. For example:

> I gave Bob a doll and Mary a gun.
> I know the problem Moscow created and failed to solve.

The former will be rejected since in *I gave Bob a doll*, *gave* is linked to both *Bob* and to *doll*. Thus, *Bob a doll* cannot be an element of a strict *and* list. In the second sentence, *Moscow* needs to connect to *failed* and *problem* must connect to *solve*, so *problems to solve* cannot be an element of a strict *and* list.

---

[10]The list of connectors allowed to do this is contained in the file `and.c` of our program. It turns out that almost all of the connector types are on this list.

This phenomenon does occur (although rarely) in formal (newspaper style) English, so we would like to solve it. The problem remains in our current system[11].

## 6.2. Encoding strict *and* lists into a link grammar

The general approach that we have taken is to add extra disjuncts to the words such that there is a linkage for the resulting sequence of words if and only if the original one contains a strict *and* list. The extra disjuncts are designed in such a way that in the resulting linkage each element of the *and* list will connect to the *and* (or to the following comma). The *and* in turn, has connectors on it so that it simulates the behavior (as far as the rest of the sentence is concerned) of one of the *and* list elements.

The extra disjuncts each contain one or more *fat connectors*. A fat connector is a connector which represents an ordered sequence of ordinary connectors. Thus, the information represented by a fat connector is analogous to that of a disjunct. A fat connector can only connect to another fat connector that represents the same sequence of connectors.

A *sub-disjunct* of a disjunct is obtained by deleting zero or more connectors from the ends of either of its two connector lists. The non-empty sub-disjuncts of:

    ((A,D)(S))

are:

    ((A,D)(S))   ((D)(S))   (()(S))   ((A,D)())   ((D)())

A fat connector will be created for each non-empty sub-disjunct of each disjunct of each word of the sequence. Each disjunct of each word will be exploded into several disjuncts. More specifically, for each sub-disjunct $d'$ of a disjunct $d$, we generate two new disjuncts: one in which the fat connector corresponding to $d'$ replaces $d'$ in $d$ and points to the left, and one with it pointing to the right.

To illustrate, we show below the disjuncts generated for ((A,D)(S)). Angle braces around (and a line under) a string indicate a fat connector representing the specified disjunct.

```
(( <(A,D)(S)> )())     (()( <(A,D)(S)> ))
((A, <(D)(S)> )())     ((A)( <(D)(S)> ))
((A,D, <()(S)> )())    ((A,D)( <()(S)> ))
(( <(A,D)()> )(S))     (()( <(A,D)()> ,S))
((A, <(D)()> )(S))     ((A)( <(D)()> ,S))
```

Now we need to describe the disjuncts on the word *and*. Let's assume (for the moment) that there is only one *and* and no commas in the sentence. In this case there will be a disjunct on *and*

---

[11]One practical (and probably effective) approach is to allow an element of an *and* list to be connected to the rest of the sentence through two different words, provided that the connectors are from among a limited set. We have deferred a detailed analysis and implementation of this idea to the future.

for each fat connector $F$. This disjunct has an $F$ pointing in each direction, and between them it has the expanded form of $F$ (its ordinary connectors). This allows *and* to accomplish its task of attaching two *and* list elements together, and to impersonate them both. For example, consider the fat connector `<(A,D)(S)>` (which, for simplicity, we will denote `F`). For this fat connector, there is one disjunct to be installed in *and*:

    ((F,A,D)(S,F))

To see how this operates, consider the second dictionary in the introduction. Here `((A,D)(S))` is a disjunct on *cat* and *dog*. The disjuncts described so far allow the following linkage:

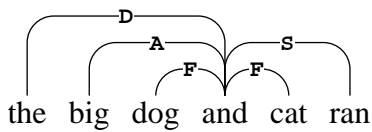

In that dictionary, *cat* also has the disjunct `((D)(S))`. If we let `G` represent the fat link `<(D)(S)>`, then the disjuncts described so far also admit the following linkage:

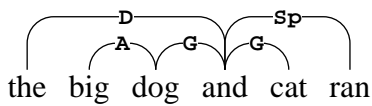

Unfortunately these disjuncts also allow

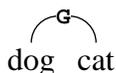

Avoiding such specious linkages is accomplished by adding some extra information to the fat connectors and modifying the matching rules for connectors. Each connector is given a *priority* of 0, 1, or 2. All normal (non-fat) connectors are given a priority of 0. The fat connectors on words are given a priority of 1, and the fat connectors on *and* are given a priority of 2 (or 1, as we'll see below). In order for two connectors to match, they must match according to the normal criteria, and their priorities must also be compatible: 0 is compatible with 0, 1 is compatible with 2, and 2 is compatible with 1. No other priorities are compatible.

Commas are easy to accommodate within this framework [12]. As with *and*, there is a disjunct on each comma for each type of fat connector. Each such disjunct consists of three fat links. For the fat connector `G` mentioned above We would create the following disjunct on the comma:

    (($G_2$)($G_1$,$G_2$))

Here the subscripts denote the priority of the corresponding connector. Note that the connector of priority 1 points to the right. This means there is exactly one way for the commas of in an *and* list to link. Here is an example.

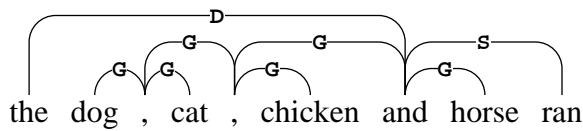

---

[12] The optional comma that can go immediately before *and* requires some extra disjuncts and connectors. A new disjunct (with one special connector to the right) is added to the comma. Each disjunct on *and* is duplicated, and one of the copies has this special connector appended to its left list.

The system we have described so far will not accommodate nesting of *and* lists. This is accomplished by adding even more disjuncts to *and*. First of all, it should clearly be possible for *and* to act in the manner of a comma, as in:

    The dog and cat and chicken and horse ran.

Thus, disjuncts like those of the comma should be added. But this is not enough. We must also add to *and* the following disjunct:

    $((F_2,F_1)(F_2))$

This disjunct allows the middle fat connector (of priority 1) to attach to its left (the comma type disjuncts allow it to attach to its right). The second *and* in the example below makes use of this feature. (The disjuncts on *or* are the same as those on *and*.)

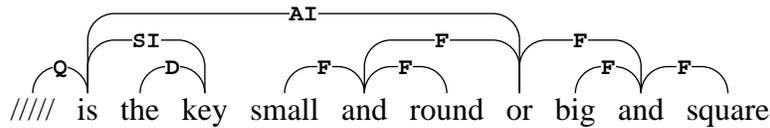

The disjuncts described so far do not allow an *and* to act as the attachment word of an *and* list element unless the fat connectors of both the outer and inner *and* lists are the same. Therefore, the following would be rejected:

    We took my heifer and goat and John's cow to the farm.

To allow this, we must treat the ordinary connectors on the *and* disjuncts in the same manner as we treated the ordinary disjuncts on words. This means substituting a fat connector for all non-empty sub-disjuncts (containing no fat connectors), and doing it in two different ways. For example let F be the fat connector `<(D,O)(M)>` and let G be the fat connector `<(O)(M)>`. Among the many disjuncts to be generated on *and* will be the following two:

    $((F_2,D)(G_1,F_2))$
    $((G_2,O)(M,G_2))$

In the following linkage the first and second of these disjuncts are used by the first and second *and* respectively.

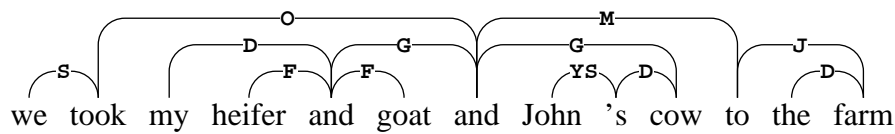

There is a subtle issue that been ignored in this discussion. How should subscripts (on the connector names) be dealt with? When two connectors with different subscripts are combined with *and*, the result must be a connector as restrictive as the most restrictive of the two. For example, consider the following dictionary:

|  |  |
|---|---|
| a | Ds+ |
| the | D+ |
| those | Dm+ |
| cats dogs | Dm- & Sp+ |
| cat dog | Ds- & Ss+ |
| ran | S- |
| run | Sp- |
| runs | Ss- |

Among the determiners above only *the* can grammatically be allowed to modify the *and* list *cats and dog*. This result is achieved in our system by the use of an *intersection* operator for connector names. The intersection of the connector names `Dm-` and `Ds-` is `D#-`. The symbol `#` matches only a `*` (which is the same as no subscript at all). Therefore `D#-` matches neither `Ds+` nor `Dmc+`, but it does match the `D+` on *the*. The result of the connector intersection operator (`D#`) is in the chosen disjunct of *and* in the phrase *cats and dog*.

This dictionary also illustrates a different problem. The system we've described so far would accept *the dog and cat runs*, while rejecting *the dog and cat run*. Both of these judgements are wrong because in English whenever two singular subjects are *anded* together they become plural. We solved this problem in our implementation by having it generate an `Sp+` connector on any disjunct on *and* that is formed by combining two subjects (`S..+` type connectors) together.

### 6.3. Some subtleties

The scheme described above can generate some rather unusual interpretations of sentences. Consider the following two linkages of the sentence *I think John and Dave ran*.

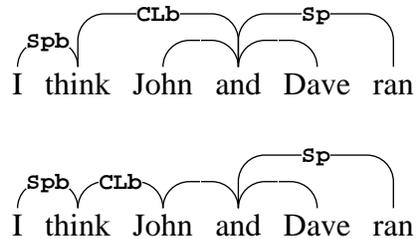

The first linkage is the natural interpretation. The second one is a combination of the following two sentences

> I think John ran.
> Dave ran.

This linkage should clearly be rejected. Intuitively, the problem with this linkage is that the same `S` link (the one between *and* and *ran*) is being used to mean something that *I think* (*John ran*) and also something that is just a fact (*Dave ran*). We have devised (but not yet implemented) a system for detecting such patterns. The idea is to use domains (see section 7). We first expand all of the *and* lists, generating a set of sentences that are represented by the original sentence. We then compute the domain structure of the resulting linkage of each sentence. Finally, the original linkage is deemed incorrect if the nesting structure of a pair of links descending from the same link (*e.g.* the `S` links in the two sentences above) do not have the same domain ancestry (are contained in the same set of domains).

An even more bizarre type of linkage sometimes emerges from from sentences with cyclic constructs. (That is, constructs that involve cycles of links.) Here is an example:

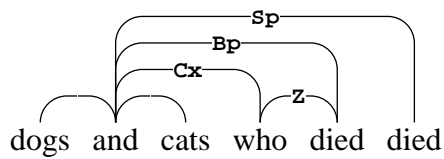

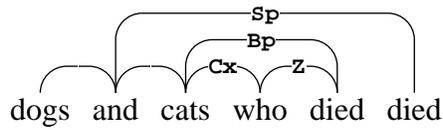
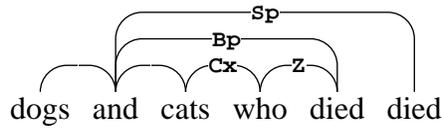

The first linkage is the natural one. The second one represents the two sentences:

> Dogs died.
> Cats who died died.

The third linkage is the same as the second one (it also represents these two sentences), but because of the cycle it is difficult to identify *cats who died* as an *and* list element. This type of linkage is eliminated by a simple post-processing step that checks that each *and* list element is attached to the rest of the sentence through the *and* only.

## 7. Post-processing

There are certain important phenomena in English that are cumbersome (perhaps impossible) to capture in a link grammar. To solve these problems, we developed a post-processing system based on a concept we call *domains*. A domain is simply a subset of the links of a linkage. The post-processing phase begins with a linkage. It classifies each link of the linkage with respect to the domains that contain it. It then checks to see if the structure of this classification satisfies several *post-processing rules*. If any of the rules are violated, then the linkage is deemed ungrammatical by our system.

The post-processing system is rather complicated. There are three different ways domains are constructed, and several dozen rules. We will not attempt to explain all of these in detail. Instead we will explain a simplified version of some aspects of the system, and give examples of two uses of its uses: non-referential *it*, and selectional restrictions. We hope that that the general approach becomes clear to the reader. A better understanding can be obtained by experimenting with the program and looking at the dictionary and the source code.

In addition to handling non-referential uses of *it* and *there*, the post-processing system is used for several other things: certain types of questions; certain uses of *whose* and *which* in relative clauses; and comparatives. Some of these capabilities are illustrated in appendix A.

### 7.1. Basic domains

In any linkage, each link has been endowed with a name. Roughly speaking this is the name of the connectors joined by the link. (Actually this name is the intersection (as defined at the end of section 6) of the two connectors joined by the link. The set of connectors that match this name is the intersection of the sets that match the connectors at the two ends of the link.) There is a set of link names called the *domain starter set*. Each link in a linkage whose name is in this set creates a domain in that linkage. The *type* of the domain (a lower-case letter) is a function of the name of the link that started it.

We can now describe how the domain $D$ started by a particular link $L$ is constructed. We think of the linkage as a graph (links are edges, words are vertices). The word at the left end of the starter link $L$ is called the *root word*. The word at the other end is called the *right word*. A link is in domain $D$ if it is in a path in the graph that starts from the right word, ends anywhere, and does not go through the root word as an intermediate vertex. (One exception to this is that the link $L$ is not part of the domain.) A depth-first search can be used to efficiently construct domains.

For example, suppose that the domain starter set is {`WA`, `CLb`}. The `WA` connector starts domains of type g and the `CLb` connector starts domains of type b. (`WA` is a special connector that allows the wall to connect to the leftmost word of any sequence of words. In the other sections of this paper we suppress the display of these links. Here we show them to clarify the exposition.)

Consider the following linkage:

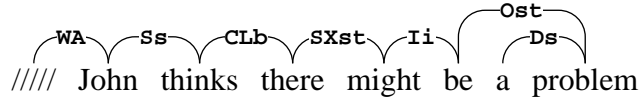
///// John thinks there might be a problem

There are seven links in this linkage. All of those to the right of the `WA` are in a `g` domain, and all of those to the right of the `CLb` are in a `b` domain. Each line of the following table shows all the information we have about a particular link. First comes set of domains containing the link, then the word at the left end of the link, then the connector at the left end of the link, then the name of the link, then the connector at the right end of the link, then the word at the right end of the link. Our program prints this information for each linkage it finds.

| Domains | Left Word | Left Con. | Link | Right Con | right word |
|---|---|---|---|---|---|
|  | ///// | WA | <--WA---> | WA | John |
| (g) | John | S | <--Ss---> | Ss | thinks |
| (g) | thinks | CLb | <--CLb--> | CL | there |
| (g) (b) | there | SXst | <--SXst-> | SX | might |
| (g) (b) | might | I | <--Ii---> | Ii | be |
| (g) (b) | be | Ost | <--Ost--> | Os | problem |
| (g) (b) | a | Ds | <--Ds---> | Ds | problem |

Now let's consider an example where post-processing can detect something non-grammatical. When *there* is used, it is attached to its object via an `SXst` type connector. One of the post-processing rules states that any domain that contains a `SXst` link must also contain an `O` type link. Consider the following example:

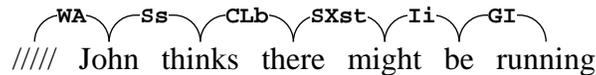
///// John thinks there might be running

```
Invalid linkage:  There rule 2 violated.
```

|  | ///// | WA | <--WA---> | WA | John |
|---|---|---|---|---|---|
| (g) | John | S | <--Ss---> | Ss | thinks |
| (g) | thinks | CLb | <--CLb--> | CL | there |
| (g) (b) | there | SXst | <--SXst-> | SX | might |
| (g) (b) | might | I | <--Ii---> | Ii | be |
| (g) (b) | be | GI | <--GI---> | GI | running |

Here the domain of type `b` contains an `SX` connector, but no `O` connector. Therefore the rule described above has been violated.

Certain link types are *restricted* in the sense that while searching for the links of a domain, if the search follows a restricted link to the left, it does not continue beyond it. A Link whose name is `B` (with any subscript) is restricted. This is illustrated in the following example. (Note that we have added another starter link `C`, which starts domains of type `r`.)

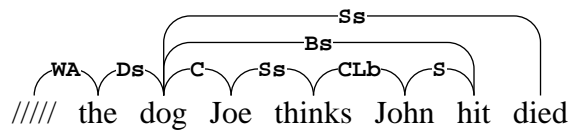
///// the dog Joe thinks John hit died

```
                      /////   WA    <--WA--->  WA   the
         (g)                  the   D     <--Ds--->  Ds   dog
         (g)                  dog   Ss    <--Ss--->  S    died
         (g) (r) (b)          dog   Bs    <--Bs--->  B    hit
         (g)                  dog   C     <--C---->  C    Joe
         (g) (r)              Joe   S     <--Ss--->  Ss   thinks
         (g) (r)              thinks CLb  <--CLb-->  CL   John
         (g) (r) (b)          John  S     <--S---->  S    hit
```

Here we see that the growth of the `b` domain has been stopped by the fact that the `Bs` link is restricted.

A domain is said to be *unbounded* if it contains a link touching a word to the left of its root word. Both the `b` and the `r` domain in the sentence above are unbounded. Domains of type `e` are started by `CLe` links. Unbounded `e` domains are not allowed. Consider the following linkage:

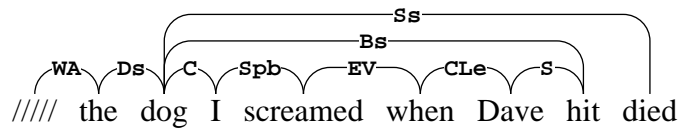

Invalid linkage:   Unbounded e domain

```
                      /////   WA    <--WA--->  WA   the
         (g)                  the   D     <--Ds--->  Ds   dog
         (g)                  dog   Ss    <--Ss--->  S    died
         (g) (r) (e)          dog   Bs    <--Bs--->  B    hit
         (g)                  dog   C     <--C---->  C    I
         (g) (r)              I     Spb   <--Spb-->  S    screamed
         (g) (r)              screamed EV  <--EV--->  EV   when
         (g) (r)              when  CLe   <--CLe-->  CL   Dave
         (g) (r) (e)          Dave  S     <--S---->  S    hit
```

Roughly speaking, domains divide a sentence into clauses. Links that are in the same clause will generally be in the same domain. It is useful in constructing the post-processing rules (as we'll see below) to be able to refer to a set of links in a given clause, but not in any subordinate clauses of that clause. To do this we introduced the concept of a *group*: a set of links which share exactly the same domain membership. Consider the following sentence:

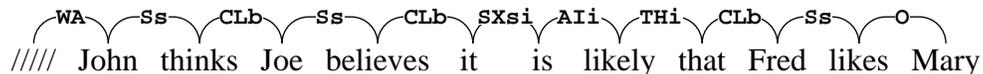

```
                      /////   WA     <--WA--->  WA   John
(g)                   John    S      <--Ss--->  Ss   thinks
(g)                   thinks  CLb    <--CLb-->  CL   Joe
(g) (b)               Joe     S      <--Ss--->  Ss   believes
(g) (b)               believes CLb   <--CLb-->  CL   it
(g) (b) (b)           it      SXsi   <--SXsi->  SXs  is
(g) (b) (b)           is      AI     <--AIi-->  AIi  likely
(g) (b) (b)           likely  THi    <--THi-->  TH   that
(g) (b) (b)           that    CLb    <--CLb-->  CL   Fred
(g) (b) (b) (b)       Fred    S      <--Ss--->  Ss   likes
(g) (b) (b) (b)       likes   O      <--O---->  O    Mary
```

Here, the `S` link between *John* and *thinks*, and the `CLb` between *thinks* and *Joe*, constitute one group. The `S` link between *Joe* and *believes*, and the `CLb` link between *believes* and *it*, constitute another, and so on.

### 7.2. Example: Judging non-referential uses of *it*

Another important application of domains is to judge sentences involving non-referential *it*. It is well known that certain adjectives can take complements such as *that* only when *it* is the subject, and then only certain verbs can be used:

> It seemed likely that John would go
> *Joe seemed likely that John would go
> *It acted likely that John would go

To enforce this, we first assign special subscripts to connectors on adjectives that can only take *that* as a complement when *it* is the subject. We also assign special subscripts to verbs that *it* can take in this situation. These subscripts have no effect on the set of linkages found; their purpose is to give information about which words are present to the post-processor. We also give special connectors to *it* and to verbs which can connect directly used with non-referential *it*. Here is a greatly simplified dictionary to illustrate the approach.

| | |
|---|---|
| act | `(Sp- or I-) & AI+` |
| seem appear | `(Sp- or SXp- or Ii-) & (AI+ or TOi+)` |
| acted | `(S- or T-) & AI+` |
| seemed appeared | `(S- or SX- or Ti-) & (AI+ or TOi+)` |
| wanted | `(S- or T-) & TO+` |
| thought | `(S- or T-) & {TH+ or CLb+}` |
| to | `TO- & I+` |
| that | `TH- & CLb+` |
| doubtful glad | `AI- & {TH+}` |
| likely | `AIi- & {THi+}` |
| he John | `S+` |
| it | `S+ or SXsi+` |

We then make the following rule: If a group contains a `THi` link then it must contain an `SXsi` link (*i.e, it* must be the subject). Furthermore, such a group cannot contain unsubscripted `T` or `I` connectors; any such connectors must be subscripted `i`. This allows the program to correctly judge the following:

> John thought it was likely that Fred would go
> *It thought John was likely that Fred would go
> It seemed to appear to be likely that John would go
> *John seemed to appear to be likely that John would go
> John seemed to appear to be doubtful that John would go
> *It seemed to want to be likely that John would go
> John seemed to appear to be likely to go

The first two of these make essential use of groups. The following diagrams should make this clear.

```
             WA     S     CLb  SXsi   AIi   THi   CLb    S      I
           /----\/----\/-----\/----\/----\/----\/----\/----\/-----\
           /////  John  thought.v  it   was  likely that Fred would  go

                        /////       WA    <--WA--->   WA    John
           (g)           John        S    <--S---->   S     thought.v
           (g)           thought.v   CLb  <--CLb-->   CL    it
           (g) (b)       it          SXsi <--SXsi->   SXs   was
           (g) (b)       was         AI   <--AIi-->   AIi   likely
           (g) (b)       likely      THi  <--THi-->   TH    that
           (g) (b)       that        CLb  <--CLb-->   CL    Fred
           (g) (b) (b)   Fred        S    <--S---->   S     would
           (g) (b) (b)   would       I    <--I---->   I     go

             WA    Ss     CLb    Ss    AIi   THi   CLb    S      I
           /----\/----\/-----\/----\/----\/----\/----\/----\/-----\
           /////  it  thought.v John  was  likely that Fred would  go
Invalid linkage:  THi rule 1 violated, THi rule 2 violated.

                        /////       WA    <--WA--->   WA    it
           (g)           it          Ss   <--Ss--->   S     thought.v
           (g)           thought.v   CLb  <--CLb-->   CL    John
           (g) (b)       John        S    <--Ss--->   Ss    was
           (g) (b)       was         AI   <--AIi-->   AIi   likely
           (g) (b)       likely      THi  <--THi-->   TH    that
           (g) (b)       that        CLb  <--CLb-->   CL    Fred
           (g) (b) (b)   Fred        S    <--S---->   S     would
           (g) (b) (b)   would       I    <--I---->   I     go
```

In the first of these linkages the non-referential form of *it* is used, and in the second the referential form is used. The violation occurs in the second linkage because the group of links whose domain ancestry is `(g) (b)` contains a `THi` link, but not a `SXsi` link. In the first linkage, this condition is satisfied.

## 7.3. Example: Selectional restrictions

Our system makes almost no observance of so-called *selectional restrictions*. That is, we do not distinguish between *person-nouns*, *thing-nouns*, *place-nouns*, and the like; nor do we distinguish between the verbs and adjectives that such nouns can take. (It seemed to be more of a semantic issue than a syntactic one.) It is a useful exercise to see how our system can be modified to enforce such restrictions.

In many cases, there are direct links between words that are logically related. A preposed adjective is always directly linked to the noun it modifies:

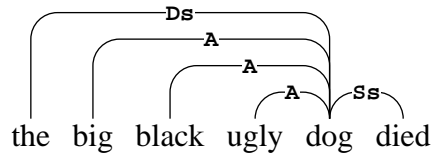

A verb is often linked to its subject, as shown four times in the following sentence:

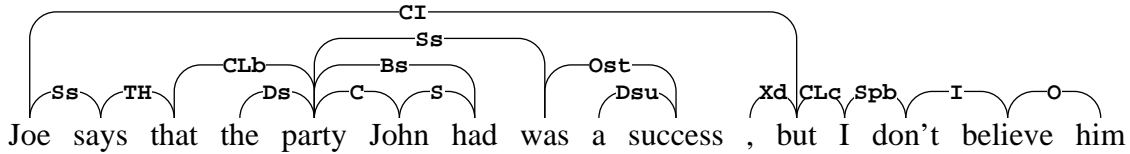

In cases where two words are linked, it is very easy to enforce selectional rules regarding their compatibility using subscripts. For example, if we wished to encode *black* as an adjective that could only modify *concrete object* nouns, we could subscript its `A+` connector as `Ac+`. Concrete object nouns could be given `Ac-` connectors. Abstract nouns such as *idea* would be given `Aa-` connectors. *The black idea died* would therefore be prevented. Of course, there can be any number of subcategorization levels.

There are also many cases where logically related words are not directly linked. For example, a noun can be separated from its main verb by auxiliaries:

            Ds   Ss   TOi  Ii   T   EV
(1)  the  dog  seems  to  have  slept  here
(2)  ?The idea seems to have slept here.

Here, the post-processing system proves useful. Generally, a group of links will contain a noun phrase, the verb of which that noun phrase is the subject, any adjectives modifying that subject, and any adverbs or prepositional phrases modifying its verb. In this case, all of the links (except the `WA` link from the wall, which is not shown) are part of one group.

We have already described post-processing rules that allow us to monitor the types of links within a group. Here we will need a rule of the form: if a group contains an `X` link as well as a `Y` link, the sentence is invalid. We might disallow sentence (2) above as follows: *Slept* would be given a `Tc-` connector (Rather than an unsubscripted `T-`, as is currently the case). *Idea* would be given an `Ssa-` (rather than an `Ss-`); *dog* would be given a `Ssc-`. We would then make a rule that if a group contained an `Ssa`, it could not contain a `Tc`. This would allow sentence (1) while rejecting sentence (2).

## 8. Choices and problems

We have attempted to endow our parser with the ability to recognize formal (newspaper style) English. As the transcripts in appendix B indicate, we have met with partial success. We can classify the failures of our system roughly into three categories:

(a) Our parser either accepts or rejects every sequence of words given to it. Since there exists no accepted definition of grammatical English, many somewhat arbitrary choices had to be made. Some English speakers may not agree with our choices.

(b) Sometimes there is a choice between (1) a simple system that makes some mistakes, and (2) a complicated system that doesn't make those mistakes. Because of our desire for elegance and simplicity, and the fact that ours is a preliminary system, we have frequently opted for (1) instead of (2).

(c) Certain constructs in grammatical English are simply unnatural in our framework.

The distinction between these three categories is actually not clear-cut. Suppose we're faced with a dilemma of type (b). Then rather than choosing option (2), we choose (1) and declare that what our system does is grammatical, moving the problem to category (a). Categories (b) and (c) are aspects of the same thing: a construct unnatural in our system can be viewed as one that would require a lot of work to implement.

The remainder of the section is a list of problems. The problems are ordered roughly from category (a) to category (c). Each problem is illustrated with sample sentences. The sample sentences are classified as clearly invalid (preceded by a "*"), of questionable validity (preceded by a "?") and valid (preceded by nothing). These are our personal judgements. Some sentences have been *marked* to indicate how our parser handles them. An "[r]" indicates that the sentence is rejected, and an "[a]" indicates acceptance. All questionable sentences are marked, as are sentences in which our parser disagrees with our judgement. (The parser performs correctly on all unmarked sentences.)

**1.** Our parser allows sentences and clauses to end with prepositions (which some might consider to be incorrect):

> The dog John screamed at died.
> Who/whom did John scream at?

Our parser also generally allows so-called *preposition prepositioning*:

> The dog at whom John screamed died.
> At whom did John scream?

**2.** Arbitrary limitations have been placed on the construction of noun phrases. We allow nouns to take a relative clause, or one or more prepositional phrases and/or participle modifiers in series (without *and*). However, we do not allow a noun to take multiple relative clauses, or a relative clause and a prepositional or participle phrase. That is, we do not allow the following:

?The man in Chicago who liked Mary died. [r]
?The man who Dave knows mentioned in the newspaper died. [r]
?The dog chasing John who Joe hit died. [r]
?The man Joe hit who likes Mary died. [r]

We do allow the following:

The dog in the park with the black nose died.
?The dog in the park bought by John died. [a]

Relative clauses using *whose*, as well as those involving prepositions (*in which*) and the like, use the M connector on nouns, unlike other relative clauses, which use the {C+} & B+ expression. These can therefore be used in series, and in combination with prepositional and participle phrases:

?The dog chasing John at whom I threw the ball died. [a]
?The dog John bought at whom I threw the ball died. [r]

We are aware that these distinctions are arbitrary, and not based on linguistic reality. However, the use of participle phrases, prepositional phrases and relative clauses in series or in combination is extremely rare. The only case seen with any frequency is prepositional phrases used in series, and these are allowed.

**3.** Our system does not handle certain kinds of quotation and paraphrase constructions frequently seen in journalistic writing:

The president would veto the bill, his spokesman announced today. [r]
Hicksville is a peaceful town, said the mayor. [r]

**4.** As pointed out in section 7, our system does not distinguish between person-nouns, thing-nouns, and place-nouns, nor does it distinguish the verbs and adjectives that such nouns can take.

?The black idea ran [a].
?The idea seems to have slept here. [a]

We made this choice because we view such distinctions as semantic rather than syntactic. The way to incorporate such *selectional restrictions* into our system is described in section 7.3.

**5.** Our parser allows present participles and question-word phrases to follow certain prepositions. This allows constructions which some might consider syntactically incorrect:

We had an argument about whether to go. [a]
?The dog about whether to go died. [a]
He is an expert on drinking beer. [a]
?The dog on drinking beer died. [a]

**6.** Sometimes present participles, *to*-[infinitive] phrases, and *that*-[clause] phrases can act as noun phrases:

Playing the piano is fun.
?Playing the piano beats going to concerts. [r]
That Joe hit John is disturbing.
To play the piano is fun.

We often find it difficult to make up our minds in such cases, and the current program's treatment of such sentences is rather erratic. However, our post-processing system isolates the subject-phrases in these sentences. For example, in the first sentence above, "playing the piano" is in one group; "is fun" is in another group. Therefore, if we could decide what kind of predicates such subject-phrases could take, this could easily be enforced using existing types of domain rules.

**7.** Our system allows two-word verbs like "walk out" and "move in". It also allows two-word transitive verbs like "tie up" and "sort out", where the two words may be separated:

> I tied up the dog.
> I tied the dog up.
> *I removed up the dog.

As shown by these sentences, our system distinguishes between verbs that can take tag-words such as "up" and "out" and those which can not. However, our system allows any tag-word-taking verb to take any tag-word. Thus we allow:

> *I tied out the dog. [a]
> *I sorted up the dog. [a]

Since there are direct connections between the verbs and their tag-words in all cases, such sentences could quite easily be prohibited using subscripts.

**8.** Adjectives appended to nouns are not allowed by our parser:

> The troops loyal to Hussein died. [r]
> ?The troops loyal died. [r]

We could easily fix this, by adding an `M-` connector to all adjectives, but we felt the cost, in terms of the number of incorrect linkages that would be found, exceeded the benefit. Furthermore, there is no good way in our system of distinguishing between the first sentence above and the second.

**9.** There are some cases besides questions where subject-verb inversion is found:

> Among the victims were several children. [r]
> Only then can we really understand the problem. [r]

We chose not to allow such sentences, since it seemed that it would result in many spurious parsings. They easily could be allowed, however, using the `SI` connectors already in place on nouns and invertible verbs. More difficult are cases like the following, found mainly in children's literature, where the inverted verb is not one usually inverted in questions:

> Out of the bushes leapt a huge tiger. [r]

**10.** The words *even* and *only*, used before noun phrases, are a problem for our system. They are treated as openers, and thus can precede any noun phrase acting as a subject:

> Even my teacher likes Joe.
> The man even John liked died.
> Even I went to the party.

This, however, results in the following spurious linkage, where *even* connects to *John*, even though it would never be understood that way:

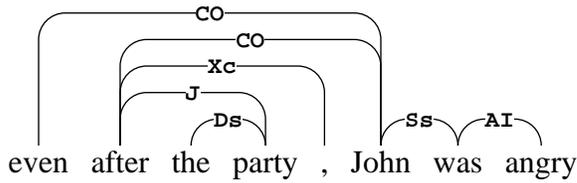

**11.** Our method of dealing with commas in noun phrases in unsatisfactory. We are unable to distinguish between the following:

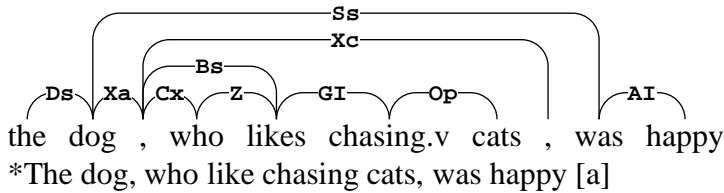
*The dog, who like chasing cats, was happy [a]

The above linkage is also troubling from a logical point of view, in that there is no direct link between *dog* and *who* (or *dog* and *likes*). (If the above sentences were modified by eliminating the commas, such links would be formed, and the second sentence would be rejected.) It is possible that commas would be better handled separately, rather than treated as elements of the dictionary. Domains (section 7) could also used to correct this problem.

**12.** Our system handles many uses of commas (see above), as well as apostrophes, both in possessives and in most contractions. However, it does does not handle any other punctuation, such as semi-colons, colons, question marks, quotation marks, and parentheses. We believe these things can all be handled under the current framework. The way in which to handle these depends on the application to which the system is to be applied. With quotation marks, for example, should words inside the quotation marks be parsed or not?

**13.** There are some limitations on the placement of adverbs and prepositional phrases modifying verbs. With transitive verbs, we allow adverbs and prepositional phrases after the direct object, not before:

> John hit the dog violently.
> ?John hit violently the dog. [r]

With verbs that take complements (such as *that*, *to* plus an infinitive, or question words) we do not allow prepositional phrases or adverbs between the verb and the complement:

> John believed deeply that Joe would come. [r]
> Joe asked me on Friday who Mary liked. [r]

**14.** Our current system does not distinguish between restrictive and non-restrictive relative clauses. That is, it allows

> The dog John hit died.
> *His dog John hit died. [a]
> The paper described by Joe was excellent.
> ?Their paper described by Joe was excellent. [a]

In general, a noun phrase taking a relative clause or a participle modifier can not take a possessive determiner such as *his*. There seem to be exceptions to this, however:

> His statements concerning abortion have been controversial.
> I love your window overlooking the park.

For this reason, we are reluctant to prohibit all possessive determiners with relative clauses and participle modifiers. We could easily do so if we chose.

**15.** Occasionally one sees a relative clause which is separated from its parent noun. Most constructions formed this way sound wrong, however. Our parser rejects all such sentences.

> Evidence was presented which disproved the claim. [r].
> ?The dog died who bit John. [r]

**16.** Some verbs seem to require a complement (such as *that* plus a clause, or *to* plus an infinitive):

> I intended to go.
> *I intended.
> I assumed he would go.
> *I assumed.

This is easily handled. Other verbs commonly take complements of this type but may take prepositional phrases instead:

> I thought he would go.
> I thought about going.
> I hoped to go.
> I hoped for a party.
> ?I hoped. [a]

The problem arises with verbs which seem to require either a complement such as *that*, or a particular preposition. Since we do not have different connectors for different prepositions, this is difficult to enforce; for example, we cannot allow *I hoped for a miracle* while barring *I hoped on Tuesday*. Of course, such special connectors could be created.

**17.** Our system makes no judgments about tense relationships between clauses. That is, clauses of differing tense may be combined in any way. This allows sentences which are nonsensical and possibly incorrect:

> ?I will arrive before he left [a].
> ?If I would have known, I have told you. [a]

Any rules for tense that might be formulated could probably be implemented using the domain framework already in place.

**18.** Our system handles idiomatic expressions with mixed success. One feature of our system is especially useful in this regard: we allow multiple-word expressions to be defined in the dictionary, in the same manner as single words. While computing the linkage, the system considers the option of using such a contiguous group of words as a single word, or as separate words (using the other definitions, if any). This allows us to handle a large number of idioms: time expressions such as "last week" and "that day"; opener expressions such as "of course" and "for example"; idiomatic prepositional phrases such as "in store" and "at hand".

In some idiomatic constructs the words of the idiom may be arbitrarily separated. Such idioms are a problem:

> I asked him a question. [r]
> (*I asked him a dog.)
> I did him a favor. [r]
> (*I did him a dog.)

Also difficult are idioms where a large number of different words may be used:

> I went/ran/drove home. [r]
> *I slept/screamed/departed home.

One common kind of idiom is the use of adverbs to modify conjunctions and prepositions:

> I like him, simply because he is nice. [r]
> We went home right after the party. [r]
> I don't like violence, even in movies. [r]

One solution to this problem is to encode such phrases – "simply because", "right after" – as idioms in the dictionary. We are currently investigating other approaches to the handling of idioms.

**19.** Our system handles many different uses of comparatives. However, there are still some problems here. For the comparative rules to work, the two halves of a comparative sentence must be in the same group. Thus sentences like the following are problematic:

> I am smarter than you think I am. [r]

Another problem is comparatives within a subject-phrase:

> Smarter men than you have been fooled by him. [r]

Also difficult are constructions like the following:

> Our program is faster than had been expected. [r]

**20.** Object preposing is not allowed:

> I don't like John, but Joe I like [r]

**21.** Our method for handling *and* has several problems, which are discussed in detail in section 6. The most severe limitation is due to our insistence on strict *and* lists. This causes the system to reject the following sentences:

> I gave Bob a doll and Mary a gun. [r]
> I know the problem Moscow created and failed to solve. [r]

**22.** Certain kinds of deletion cannot be handled by our parser:

> John hit the dog, Fred the cat. [r]
> If you can't do it, you should find someone who can. [r]

## 9. Remarks

Almost all of the problems we have noticed with our current system (section 8) can be solved within the same framework. We have made steady progress in the quality of our dictionary and post-processing system, and there is no reason to believe that this could not be continued. It therefore appears that our system could, without any major conceptual breakthroughs, be turned into a useful grammar checker.

However, it appears to be possible to make use of our ideas in a variety of other natural language processing applications. Below are a few speculative reasons for our optimism.

**Uses of linkages**

Our current system only uses the linkages it computes for checking if the sentence is grammatical. Are these linkages useful for doing anything else? We have reason to believe that they are.

First of all, after seeing several linkages produced by our system, one soon gains intuition about how the meaning of a sentence is related to the structure of the linkage. This, and the connection to dependency structures [8, 6] are encouraging signs that the semantics of a sentence can be analyzed by means of the linkages our system produces.

There are also some ways in which our linkages may be more useful than the parse trees produced by a context-free based system. Joshi points out [7, p. 1246] that an advantage of a word-based system (such as ours) over context-free grammar based systems is that in a word-based system it is easier to gather meaningful statistical data about the relationships between words. Suppose, for example, that the frequency with which pairs of words are linked with a particular type of link is accumulated. Then, when the system is trying to decide which linkage (among several choices) is the correct one, it can call on the statistical data to determine which one is the most likely.

**Uses of the dictionary**

In many language processing applications, it is useful to create specialized grammars. For example, in a speech understanding system it is useful to be able to restrict the grammar (and the vocabulary) to a small subset of English. This restricts the number of grammatical strings, and simplifies the task of choosing the sentence that best fits the speech signal. It is always easy to restrict the vocabulary, but it is not always easy to control the grammar.

It appears that the structure of our dictionary is well-suited to be automatically simplified. For example, if it was decided that relative clauses were irrelevant for some application, then every occurrence of B, C, and Z could be removed from the dictionary. The same idea applies to questions, conjunctions, and many other features of our dictionary.

The grammar can also be made more forgiving. This could be done either by deleting subscripts from some subset of the connectors, or by modifying the connector matching algorithm to allow mismatches. Such changes might be useful for analyzing conversational English. For teaching English as a foreign language, if the system encountered a non-grammatical sentence, it could resort to a less restrictive matching rule. This might allow the system to identify the problem. (Actually, our current system allows costs to be associated with particular links. This feature is not described in this paper.)

**Uses of the algorithms**

There is another connection between context-free grammars and link grammars. Consider a context-free grammar in *single-terminal form*. This means that every production has exactly one terminal on the right-hand side. (Greibach normal form is a special case of this. We do not know how much blow-up in the size of the grammar occurs in this transformation.) We can make this into a link grammar as follows. Each non-terminal corresponds to a connector pointing to the left or right (two non-terminals per connector type). The disjuncts for a word correspond to each of the productions that contain that word on the right. The wall has a *start* connector pointing to the right. Now, we've created a link grammar from the context free grammar. All of the powerful pruning techniques we've developed will work here. (Note that we'll have to modify our algorithm so that cyclic linkages are not allowed. This is easy.)

## A   Examples

We have described (section 3) some basic uses of different categories of words: nouns, verbs, etc. Our on-line dictionary allows a number of other uses of these categories, and also makes many distinctions within categories. The sentences below illustrate many of these uses and distinctions. All sentences marked with "*" are rejected by the on-line dictionary; all unmarked sentences are accepted.

**Nouns**
>The fact/*event that he smiled at me gives me hope
>I still remember the day/*room I kissed him
>But my efforts/*presents to win his heart have failed
>Failure/*Absence to comply may result in dismissal
>She is the kind/*character of person who would do that
>An income tax increase may be necessary
>*A tax on income increase may be necessary
>A tax on income may be necessary
>Last week/*dog I saw a great movie
>The party that night/*house was a big success
>John Stuart Mill is an important author
>The Richard Milhous Nixon Library has been a big success
>The mystery of the Nixon tapes was never solved
>*The mystery of the tapes Nixon was never solved
>John, who is an expert on dogs, helped me choose one
>*John who is an expert on dogs helped me choose one
>John, an expert on dogs, helped me choose one
>The dog that we eventually bought was very expensive
>*The dog, that we eventually bought, was very expensive
>*The dog, we eventually bought, was very expensive
>Have you ever seen the/*a Pacific
>Mary's family is renovating their kitchen
>*A woman I know's family is renovating their kitchen
>The boys'/*boys's bedrooms will be enlarged
>My uncle's mother's cousin is visiting us
>*John's my cousin is visiting us

**Determiners and Pronouns**
>Many people/*person were angered by the hearings
>Many were angered by the hearings
>My many/*some female friends were angered by the hearings
>Many who initially supported Thomas later changed their minds
>The stupidity of the senators annoyed all/*many my friends
>I'm looking for someone special

    *I'm looking for a man special
    Anyone who thinks this will work is crazy
    Their program is better than ours
    Those that want to come can come
    About 7 million people live in New York
    About 7000000/*many people live in New York
    The city contains over one hundred million billion brain cells
    Only 5 million are in use at any time

**Question Words**
    What/*Who John did really annoyed me
    Whoever/*Who wrote this program didn't know what they were doing
    Invite Emily and whoever else you want to invite
    The dog which/*what Mary bought is really ugly
    I wonder whether we should go
    *Whether should we go
    How do you operate this machine
    How fast is the program
    How certain/*tired are you that John is coming
    How likely is it/*John that he will come
    How certain does he seem to be that John is coming
    Do you know how to operate this machine
    How/*Very/*Where quickly did you find the house
    How much/*many money do you earn
    I'll show you the house where/*which I met your mother
    This is the man whose/*which dog I bought
    I wonder where Susan is/*hit
    The dog the owner/*owners of which likes Mary died
    The dog the owner of which John hit died
    Today I saw the dog the owner of which likes John
    *The dog John hit the owner of which died
    *The dog the owner of the dog hit John died
    *The owner of which died
    *The dog the owner the sister of which hit died
    *The dog the owner Mary thinks hit the sister of which died
    The books the size of the covers of which is/*are regulated by the government are here

**Conjunctions**
    The woman we saw when/*but we went to Paris is here
    You should see a play while/*after in London
    I left the party after/*because/*despite seeing John there
    Because/*Therefore I didn't see John, I left
    I left, therefore I didn't see John
    But/*After I really wanted to see him

**Prepositions**
    I have doubts about/*behind going to the movie
    From your description, I don't think I would enjoy it
    We had an argument over/*at whether it was a good movie
    Because of the rain, we decided not to go
    They're having a party in front of the building
    The fellow with whom I play tennis is here
    The fellow I play tennis with is here
    *The fellow whom I play tennis is here
    *The fellow with whom I play tennis with is here
    With whom did you play tennis
    Who did you play tennis with

**Adjectives**
    You are lucky/*stupid that there is no exam today
    You are lucky/*right I am here
    This is something we should be happy about
    *This is something we should be happy about it
    *The happy about it woman kissed her husband
    Is he sure/*correct how to find the house
    You have a right to be proud/*happy of your achievement

**Adverbs**
    She is/*knows apparently an expert on dogs
    Mary suddenly/just left the room
    Suddenly/*Just, Mary left the room
    He told them about the accident immediately/*presumably
    He is very careful about her work
    He works very carefully
    *He very works carefully
    Is the program fast enough/*too for you
    Is the program too/*enough fast for you

*It* **and** *There*
    There is/*are a dog in the park
    *There is running
    Does there seem/*want to be a dog in the park?
    There seems to appear to have been likely to be a problem
    *There seems to appear to have been likely to be problems
    *There seems to appear to have been likely to be stupid
    There was an attempt to kill John
    *The man there was an attempt to kill died
    There was a problem, but we solved it

It/*Joe is likely that John died
It/*Joe is clear who killed John
It may not be possible to fix the problem
Mary may not be possible to fix the problem
Flowers are red to attract bees
Mary is easy/*black to hit
It is important to fix the problem
Mary is important to fix the problem
The man it/*Joe is likely that John hit died
Does it seem likely that John will come
Does John act glad that Joe came
*Does it act likely that Joe came
It/*Joe doesn't matter what John does
I want it to be possible to use the program
I want/*asked it to be obvious how to use the program
*I want John to be obvious how to use the program
I want Joe to be easy to hit

**Comparatives**
Our program works more elegantly than yours
Ours works more elegantly than yours does
*Ours works more elegant than yours
Our program works more elegantly than efficiently
Our program is more elegant than efficient
Our program works better than yours
We do this more for pleasure than for money
He is more likely to go than to stay
*He is more likely than to stay
*He is more black to go than to stay
He is more likely to go than he is to stay
He is more likely to go than John is
It is more likely that Emily died than that Fred died
It/*John is more likely that Emily died than it is that Fred died
*It is more likely that Emily died than John is that Fred died
It/*John is easier to ignore the problem than to solve it
It is easier to ignore the problem than it is to solve it
Our program is easier to use than to understand
*Our program is easier to use it than to understand
I am more happy now than I was/*earned in college
He is more a teacher than a scholar
I make more money in a month than John makes/*dies in a year
*I hit more the dog than the cat
I have more money than John has time
*I have more dogs than John has five cats
*I have more money than John has a dog

I am as intelligent as John
I earn as much money as John does
*I am as intelligent as John does
I earn as much money in a month as/*than John earns in a year

*And*, **etc.**
I went to the store and got a gallon of milk
I went and got a gallon of milk
*I got and went a gallon of milk
I got a gallon of milk and some eggs
I went to the store, got a gallon of milk, and returned the eggs
*I went to the store, got a gallon of milk, and some eggs
I went to the store and got a gallon of milk and some eggs
Mary, Joe and Louise are coming to the party
Neither Mary nor/*and Louise's friend are coming to the party
I am ready and eager to go to the party
She handled it skillfully and with compassion
The animal either went under the wall, or climbed over it
I told him that I hated her and never wanted to see her again
She told me why she was here and what she was doing
*She told me why she was here and that she hated me
Although she likes and respects me, she says she needs some privacy
Your house and garden are very attractive
Give me your money or your life
I am in New York and I would like to see you
This is not the woman we know and love
*This is not the woman we know and love John
The coverage on TV and on the radio has been terrible
*The coverage I have seen and on TV has been terrible
She and I are/*is/*am friends
My dogs and cat are/*is in the house
*A/*These dogs and cat are in the house
Ann's and my/John's father's house is huge
Ann and my father's/*father house is huge
Ann's/*Ann and my house is huge
My dog and cat, and John's horse were at the farm
*My dog, cat, and John's horse were at the farm

## B  Transcripts

We applied our program to two short columns from the Op-Ed page of the New York Times in their entirety (each contains 10 sentences). The result of this experiment are contained in this section.

A remark is in order about the method used to make these transcripts. When looking at a transcript such as ours, a skeptical reader might (justifiably) ask: How much was the system modified specifically in order to handle these sentences? We made two types of modifications: adding words to existing categories, and changing the definitions of existing words. All of the changes we made satisfy the following criterion: they are general good additions to the system, not specifically tailored to handle these sentences. For example, we modified the definition of *people* to allow its use as singular noun, as in *Americans are a war-like people*. Another example is the word *tied*. We could have modified it so that it could act as an adjective. This would have allowed sentence 5 of Transcript 1 to be accepted without any changes. We didn't make this change, because in our view it's not correct[13].

For each sentence, we first show the sentence exactly as printed in the Times. If the sentence had to be modified to be processed by our program, the modified version of the sentence is also shown (marked with ">"). Following this is a diagram showing the first, or *preferred* linkage generated by our program. (The program rates each linkage based on certain heuristics – total link length, eveness of the and lists, etc. The linkages are sorted and numbered according to this rating. This information is also displayed in parentheses.) If the preferred linkage is not the correct one, we include a display of the correct linkage. After all of this is a note (marked with a "*") explaining the differences (if any) between the printed version and the version given to the program, and the mistakes (if any) the program made in choosing the correct linkage.

In the first article the program accepted 4 of the 10 sentences exactly as they were printed. The remaining 6 were accepted with slight modifications. In 5 out of 10, the preferred linkage was correct.

In the second article 7 out of 10 sentences were parsed exactly as printed (one sentence had to be broken into two, and two others required slight changes). In 6 of the resulting 11 sentences, the preferred linkage was correct.

The computing time required to process each of these sentences is shown in appendix C.

---

[13]The directory `/usr/sleator/public/grammar.transcript` contains the version of the program used to make this transcript.

## B1.  Transcript 1

**Democracy's Chance in Russia**
by Blair Ruble
New York Times, August 29, 1991, p. A29.

**1. Boris Yeltsin has demonstrated extraordinary courage in the defense of democratic values.**

3 ways to parse (3 linkages)

linkage 1 (p.p. violations: 0, disjunct cost: 0, and cost: 0, link cost: 3)

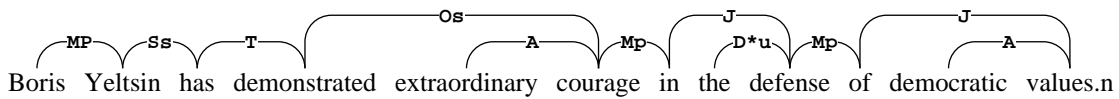

linkage 2 (p.p. violations: 0, disjunct cost: 0, and cost: 0, link cost: 5)

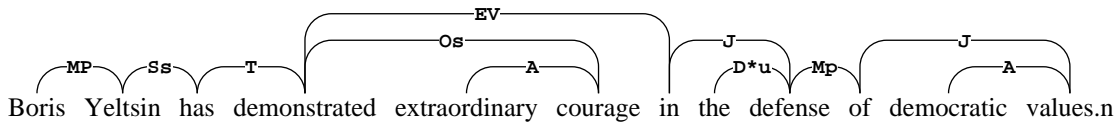

* Parsed exactly as printed.  The preferred linkage is wrong on one ambiguity.  The program guessed that *in* modifies *courage* rather than *demonstrated*.  Linkage #2 is correct.

**2. Yet his more recent claims of expanding his Russian republic's lands understandably give his neighbors pause.**

3 ways to parse (3 linkages)

linkage 1 (p.p. violations: 0, disjunct cost: 0, and cost: 0, link cost: 21)

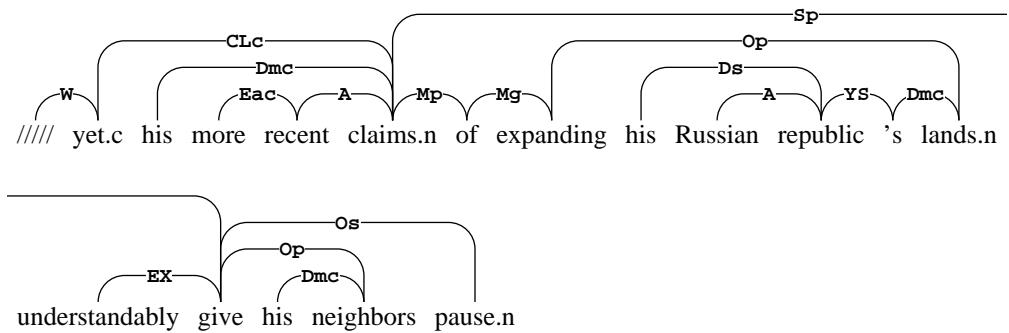

\* Parsed exactly as printed. The preferred linkage is correct.

### 3. Like the Soviet Union itself, the republic is an artificial construct, embracing scores of nationalities and dozens of homelands.

> Like the Soviet Union, the republic is an artificial construct, embracing scores of nationalities and dozens of homelands

5 ways to parse (5 linkages)

linkage 1 (p.p. violations: 0, disjunct cost: 0, and cost: 0, link cost: 19)

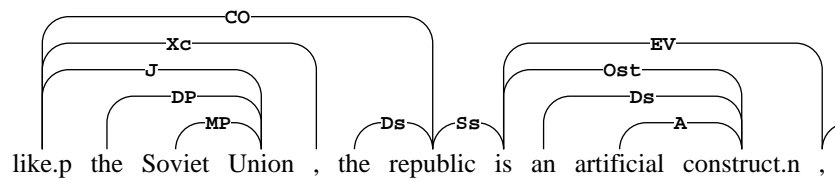

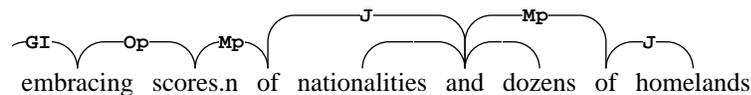

linkage 3 (p.p. violations: 0, disjunct cost: 0, and cost: 0, link cost: 22)

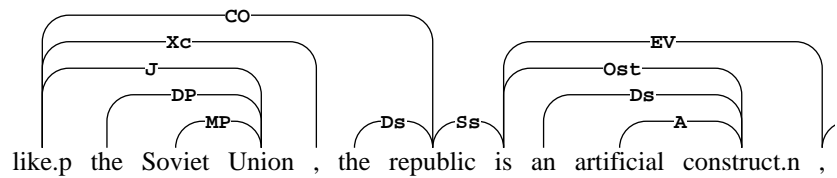

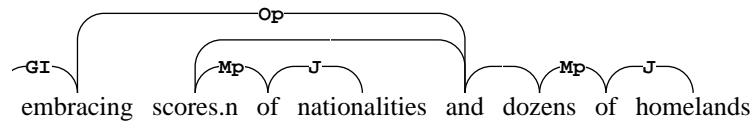

\* The program does not allow *the Soviet Union itself*; *the Soviet Union* is substituted. The program guesses wrong on one ambiguity, putting *nationalities* and *dozens* in an *and*-list instead of *scores* and *dozens*. Linkage #3 is correct.

**4. If Russian leaders can rally support for 12 million Russians living in the Ukraine, Ukrainian leaders can bemoan the fate of some six to eight million of their compatriots in Russia.**

> If Russian leaders can rally support for 12 million Russian people living in the Ukraine, Ukrainian leaders can bemoan the fate of six million of their compatriots in Russia

105 ways to parse (105 linkages)

linkage 1 (p.p. violations: 0, disjunct cost: 0, and cost: 0, link cost: 39)

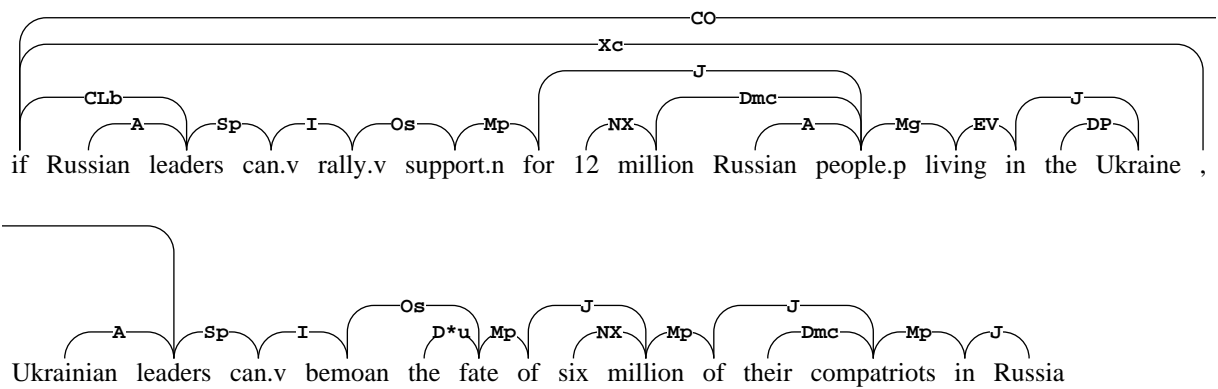

\* Two changes are necessary here. 1) The phrase *Russians living in the Ukraine* is a problem because the program treats *Russians* as a proper noun and thus does not allow it to take a participle phrase. Therefore we substitute *Russian people*. 2) The phrase *six to eight million* is an idiom not currently in our system, as is the phrase *some six million*. We substitute *six million*. The preferred linkage is correct.

**5. Danger exists in having ethnic identity becoming too closely tied to republic borders.**

> Danger exists in having ethnic identity be too closely tied to republic borders

3 ways to parse (3 linkages)

linkage 1 (p.p. violations: 0, disjunct cost: 0, and cost: 0, link cost: 6)

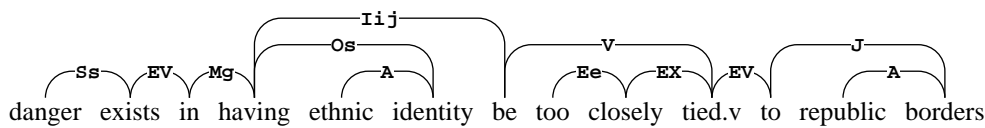

\* One change is necessary. We allow *become* to take an adjective, but not a past participle, and we do not consider *tied* to be an adjective. Thus we substitute *be* for *become*. The preferred linkage is correct.

**6. Mechanisms must be developed to encourage the emergence of multiethnic societies and states, and to provide for peaceful adjudication of border disputes.**

> Mechanisms must be developed to encourage the emergence of multiethnic societies and states, and to provide peaceful adjudication of border disputes.

5 ways to parse (5 linkages)

linkage 1 (p.p. violations: 0, disjunct cost: 3, and cost: 2, link cost: 25)

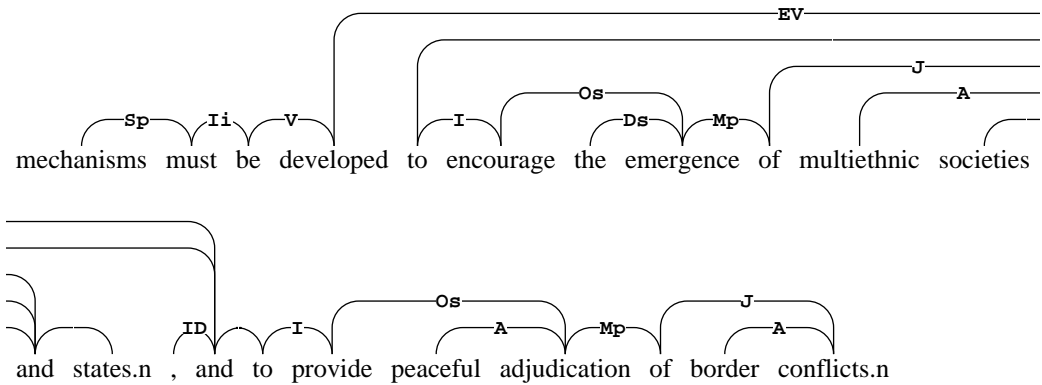

\* One change is necessary. *provide for* is an idiom not currently installed; we substitute *provide*. The preferred linkage is correct.

**7. Whatever his recent statements, Mr Yeltsin has handled the conflicts with considerable sensitivity, piecing together intricate republic-to-republic agreements.**

> Despite his recent statements, Mr Yeltsin has handled the conflicts with considerable sensitivity, piecing together intricate republic-to-republic agreements

2 ways to parse (2 linkages)

linkage 1 (p.p. violations: 0, disjunct cost: 0, and cost: 0, link cost: 22)

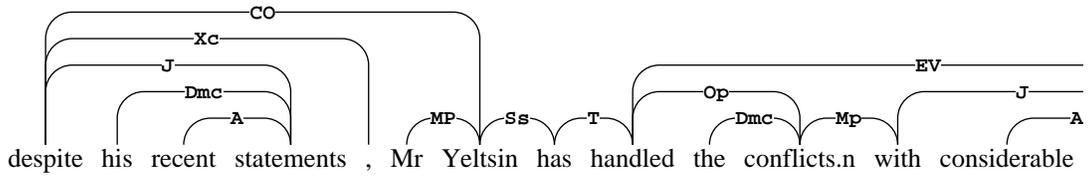

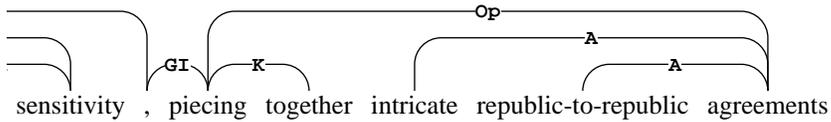

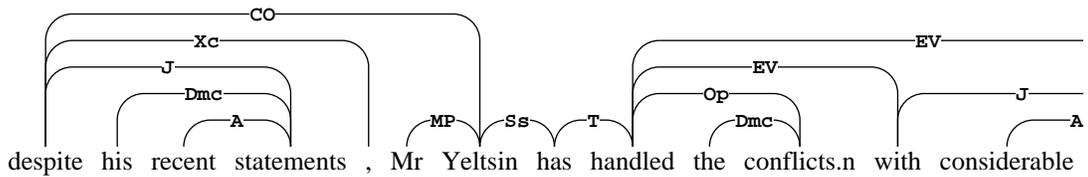

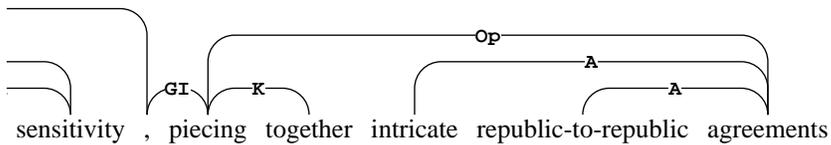

\* One change is necessary: *Whatever his recent statements* is problematic. We substitute *Despite...*. The preferred linkage is wrong on one ambiguity: it guesses that *with* modifies *conflicts* rather than *handles*. Linkage #2 is correct.

## 8. For their part, Russians have demonstrated surprisingly little bloodlust in light of their privations and dislocations.

21 ways to parse (21 linkages)

linkage 1 (p.p. violations: 0, disjunct cost: 1, and cost: 0, link cost: 11)

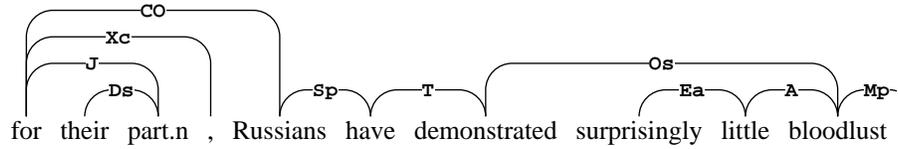
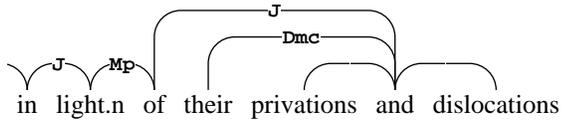

linkage 9 (p.p. violations: 0, disjunct cost: 1, and cost: 0, link cost: 14)

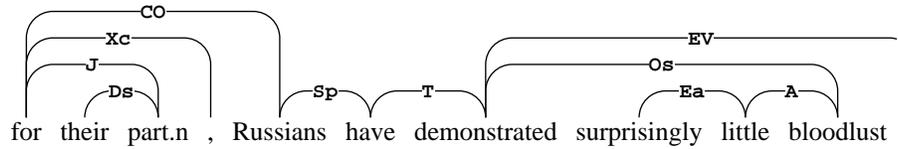
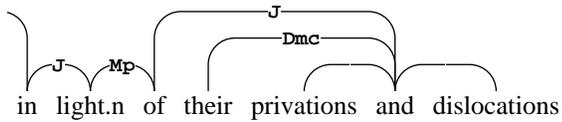

\* Parsed exactly as printed. The preferred linkage is wrong in one place: it thinks that *in light of* modifies *bloodlust* instead of *demonstrated*. Linkage #9 is correct.

**9. The victorious resistance to the coup is yet another manifestation of a commitment to democratic values that has been apparent in demonstrations, campaigns and kitchen table discussions for some time.**

52 ways to parse (52 linkages)

linkage 1 (p.p. violations: 0, disjunct cost: 0, and cost: 2, link cost: 21)

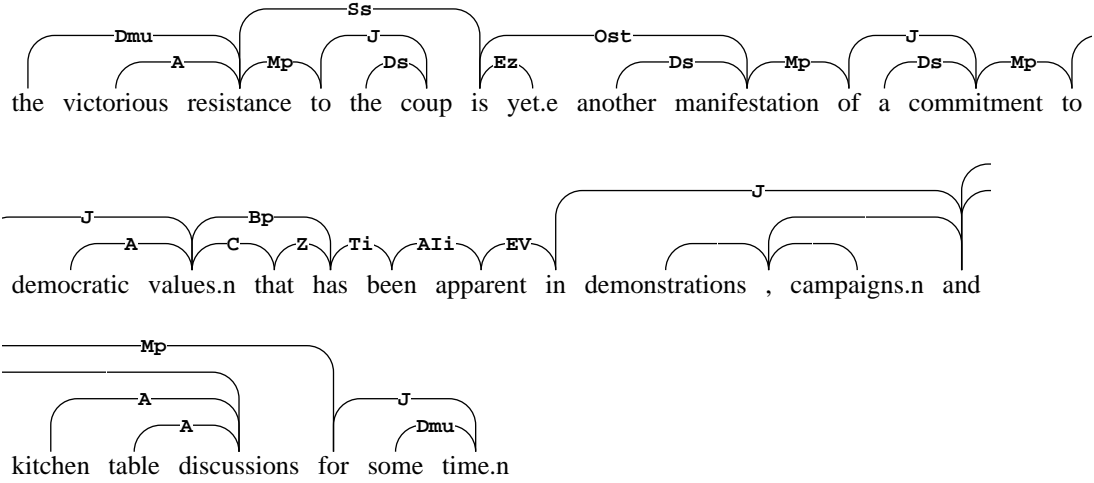

linkage 3 (p.p. violations: 0, disjunct cost: 0, and cost: 2, link cost: 26)

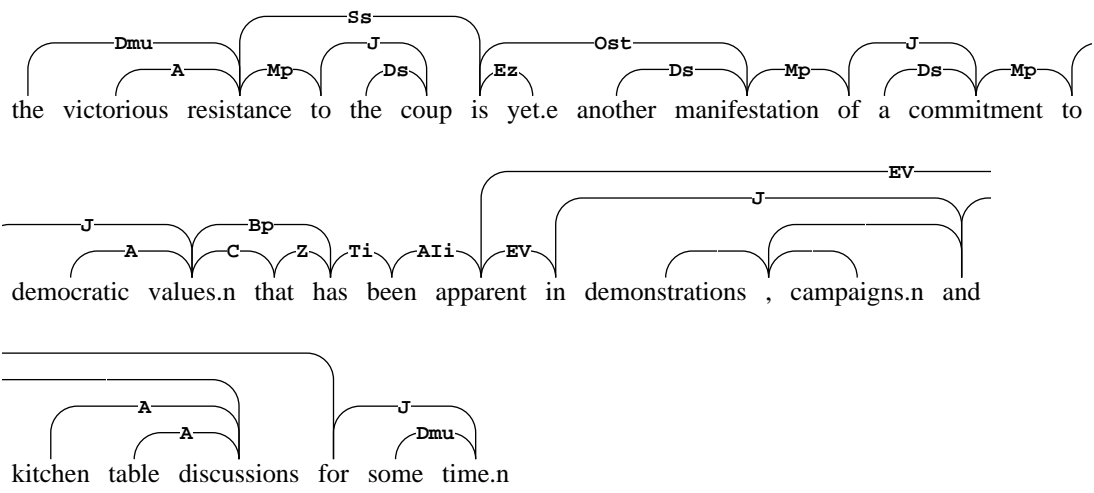

\* Parsed exactly as printed. The preferred linkage is wrong in one place. It guesses that *for* modifies *demonstrations* rather than *apparent*. Linkage #3 is correct.

### 10. A new Russian society has come into being.

> A new Russian society has come into existence

unique parse (p.p. violations: 0, disjunct cost: 0, and cost: 0, link cost: 3)

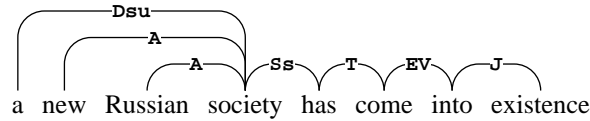

a new Russian society has come into existence

\* One change is necessary. The idiom *come into being* is problematic; we substitute *come into existence*. A unique linkage is found, which is correct.

## B2. Transcript 2

### An Islamic Empire?
by Martha Brill Olcott
The New York Times, August 29, 1991, p. A29.

**1. The coup came at a bad time for Central Asia's leaders.**

2 ways to parse (2 linkages)

linkage 1 (p.p. violations: 0, disjunct cost: 0, and cost: 0, link cost: 6)

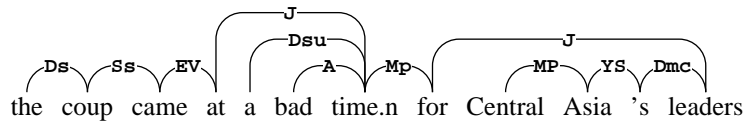

the coup came at a bad time.n for Central Asia 's leaders

linkage 2 (p.p. violations: 0, disjunct cost: 0, and cost: 0, link cost: 10)

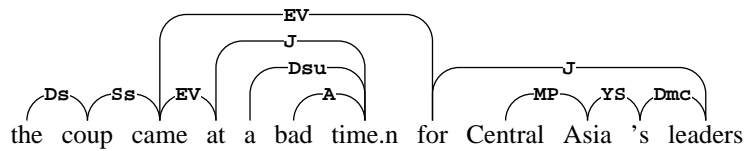

the coup came at a bad time.n for Central Asia 's leaders

\* Parsed exactly as printed. The preferred linkage guesses that *for* modified *time* rather than *came*. Linkage #2 is correct. (One could argue that the preferred linkage is better.)

**2. They must now demand independence or risk political defeat, despite the region's overpopulation and underdevelopment that would be better helped by a more gradual transition.**

135 ways to parse (189 linkages)

linkage 1 (p.p. violations: 0, disjunct cost: 0, and cost: 0, link cost: 23)

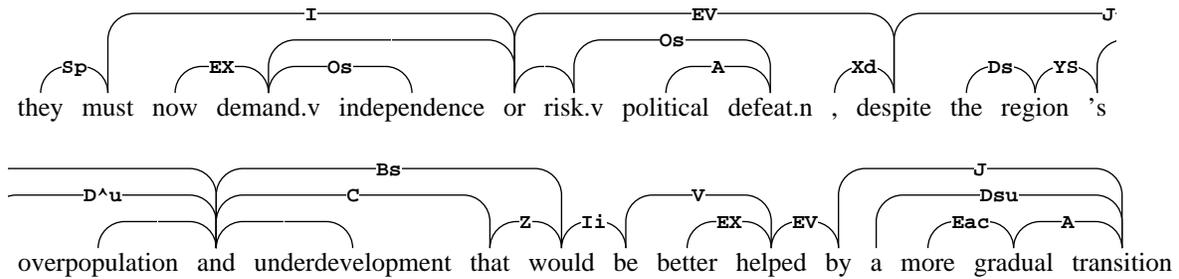

linkage 11 (p.p. violations: 0, disjunct cost: 0, and cost: 1, link cost: 25)

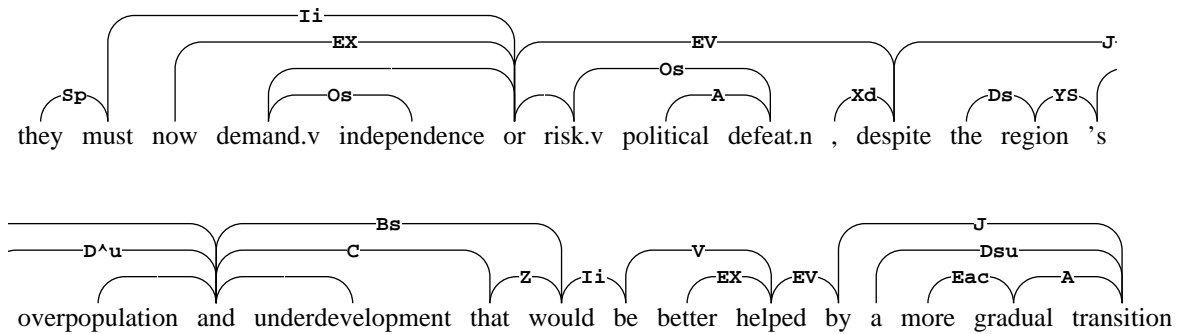

\* Parsed exactly as printed. The only difference between the preferred linkage and the correct one (#11) is in the placement of *now* inside of an and list (for the word *or*) instead of outside of it.

### 3. These artificial states were created by Soviet administrators to keep the region's Muslims from unifying.

> These artificial states were created by Soviet administrators to keep the region's Muslim people from unifying

6 ways to parse (6 linkages)

linkage 1 (p.p. violations: 0, disjunct cost: 1, and cost: 0, link cost: 10)

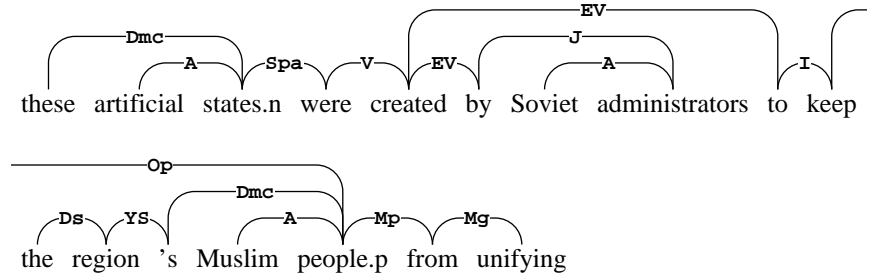

linkage 4 (p.p. violations: 0, disjunct cost: 1, and cost: 0, link cost: 15)

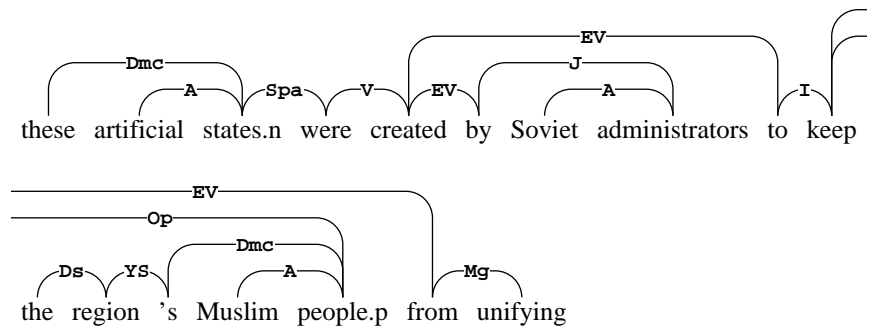

\* One change is necessary here. *Muslims* is treated as a proper noun and can not take modifiers like *the region's*; we substitute *Muslim people*. The preferred linkage guesses that *from* modifies *people* rather than *keep*. Linkage #2 is correct.

**4. Now the Central Asians must decide if they are a single people or five nationalities.**

2 ways to parse (2 linkages)

linkage 1 (p.p. violations: 0, disjunct cost: 0, and cost: 1, link cost: 8)

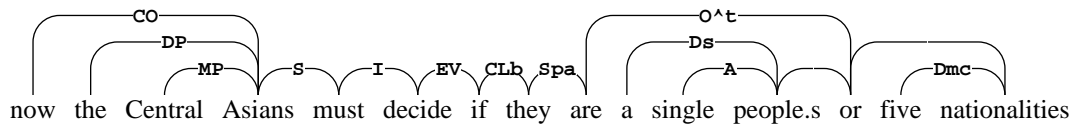

linkage 2 (p.p. violations: 0, disjunct cost: 0, and cost: 1, link cost: 8)

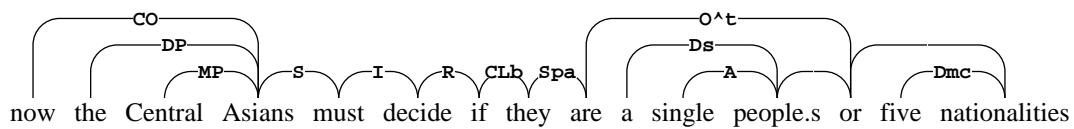

\* Parsed exactly as printed. The preferred linkage makes a subtle wrong guess. It guesses that *if* is acting as a conjunction here - hence the `EV` connector between *decide* and *if* - rather than as an indirect question word, in which case it would connect to *decide* with an `R` connector. Linkage #2 is correct.

### 5. Uzbekistan is a viable independent state, but would probably not be content with its present boundaries: Uzbeks spill over into Kazakhstan, Kirghizia, Tadzhikistan and Turkmenistan.

> Uzbekistan is a viable independent state, but it would probably not be content with its present boundaries

unique parse (2 linkages)

linkage 1 (p.p. violations: 0, disjunct cost: 0, and cost: 0, link cost: 18)

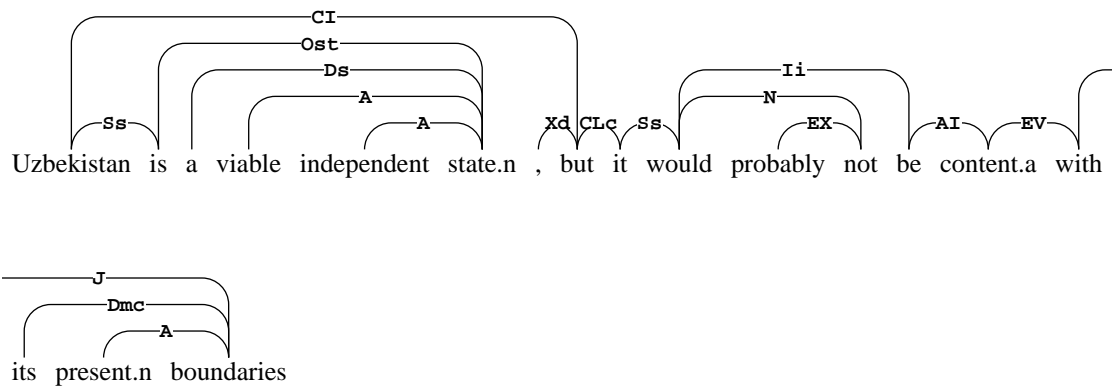

> Uzbeks spill over into Kazakhstan, Kirghizia, Tadzhikistan and Turkmenistan

unique parse

(p.p. violations: 0, disjunct cost: 0, and cost: 0, link cost: 8)

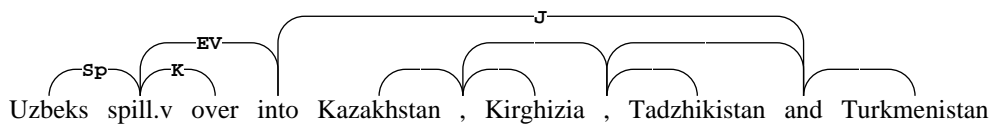

\* Our parser cannot yet handle semi-colons, colons, and hyphens; therefore we split this sentence into two. In the first half of sentence, one change is necessary. Our parser can not handle this use of *but*. Instead of *Uzbekistan is a viable independent state, but would...*, we substitute *Uzbekistan is a viable independent state, but IT would...*. The preferred linkage is correct. The second half of the sentence is parsed exactly as printed. A unique linkage is found, which is correct.

**6. Kazakhstan is wealthy with natural resources, but it couldn't exist independently, as Kazakhs make up only 40 percent of the republic's population.**

2 ways to parse (4 linkages)

linkage 1 (p.p. violations: 0, disjunct cost: 0, and cost: 0, link cost: 15)

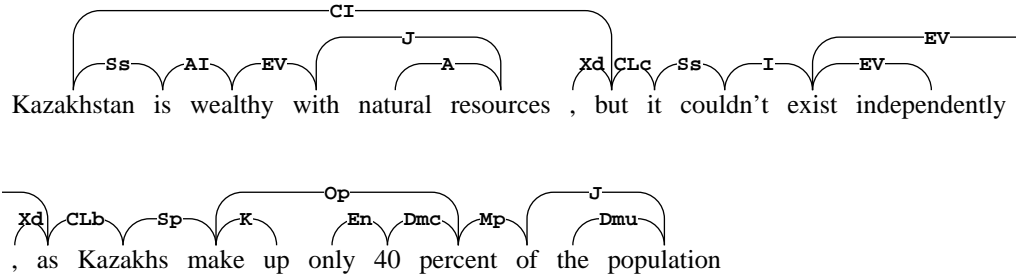

* Parsed exactly as printed. The preferred linkage is correct.

**7. Russians in northern Kazakhstan would secede and seek annexation by Siberia, though this week's supposed tempest between Kazakhs and Yeltsin supporters was greatly overblown.**

> Russian people in Northern Kazakhstan would secede and seek annexation by Siberia, though this week's tempest between Kazakhs and Yeltsin supporters was greatly overblown

12 ways to parse (12 linkages)

linkage 1 (p.p. violations: 0, disjunct cost: 0, and cost: 1, link cost: 25)

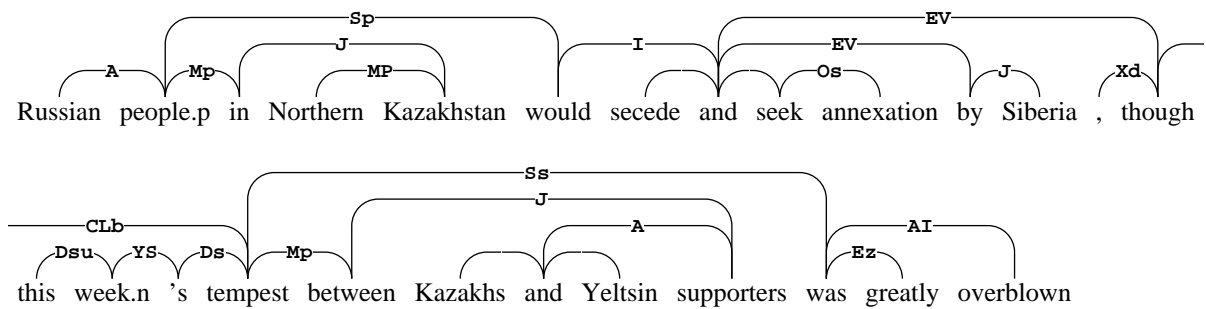

linkage 4 (p.p. violations: 0, disjunct cost: 0, and cost: 4, link cost: 21)

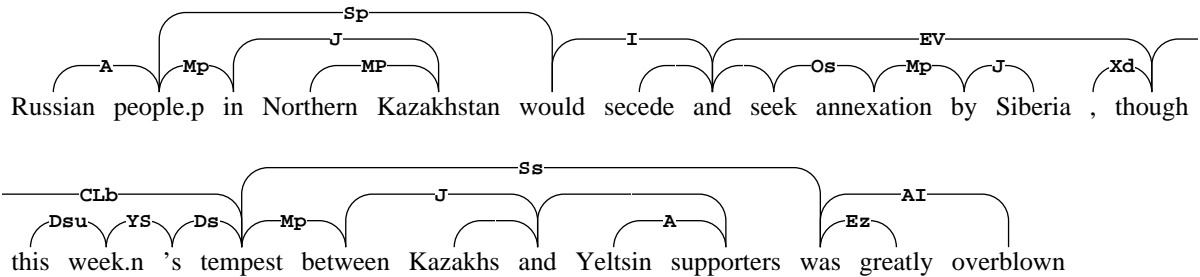

\* Three changes are necessary here. *Russians* is treated as a proper noun and cannot take a prepositional phrase; we substitute *Russian people*. *Kazakhstan* is a proper noun and cannot take an adjective; we capitalize *northern*, making this simply a two-word proper noun phrase. The parser also cannot handle *supposed* as a modifier preceding a noun; we omit this word. The preferred linkage guesses wrong twice. It guesses that *by* modifies *secede and seek* rather than *annexation*; and it forms an *and*-list using *Kazakhs* and *Yeltsin* (both modifying *supporters*), rather than *Kazakhs* and *Yeltsin supporters*.

### 8. If Kazakhstan splits apart, it will be hard for the other republics to resist Uzbek expansion.

4 ways to parse (8 linkages)

linkage 1 (p.p. violations: 0, disjunct cost: 0, and cost: 0, link cost: 15)

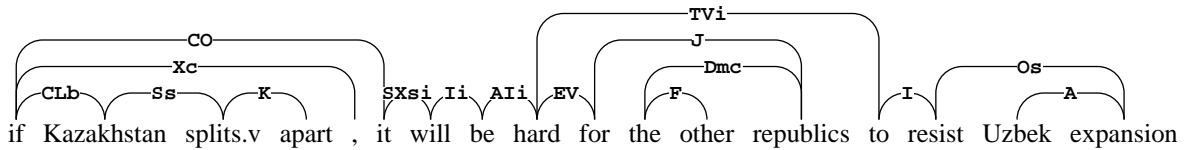

\* Parsed exactly as printed; the preferred linkage is correct. (The SX link between *it* and *is* indicates that the parser recognizes this as a non-referential use of *it*.)

### 9. Uzbekistan is the cradle of the region's Islami civilization, and its leaders dream of a Central Asian empire.

12 ways to parse (12 linkages)

linkage 1 (p.p. violations: 0, disjunct cost: 0, and cost: 0, link cost: 22)

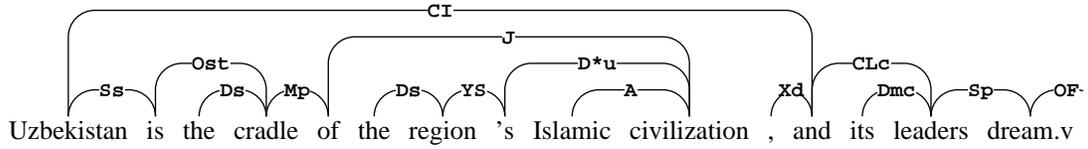

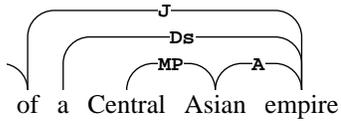

\* Parsed exactly as printed. The preferred linkage is correct. (The `OF` link between *dream* and *of* is due to the fact that *of* is not normally allowed to modify verbs, that is, it has no `EV` connector; verbs which can take *of*, like *dream*, *think*, and *accuse*, are given `OF` connectors.)

**10. Now this vision is secular, but deteriorating economies will favor Islamic radicalism.**

unique parse

(p.p. violations: 0, disjunct cost: 0, and cost: 0, link cost: 6)

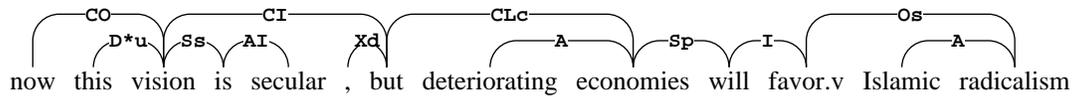

\* Parsed exactly as printed. A unique linkage is found, which is correct.

## C  Running times of transcript sentences

The following table shows the amount of time used by each phase of our program on each of the 21 sentences of the transcript (appendix B). The version of the program clocked here is in `/usr/sleator/public/grammar.transcript`. The program was running on a Decstation 3100. The meaning of each column, and the sentence corresponding to each row are listed after the table.

|      | Build | Prune | Dup  | And  | Prune2 | Power | Count | Post | Total |
|------|-------|-------|------|------|--------|-------|-------|------|-------|
| 1-1  | 0.54  | 0.17  | 0.01 | 0.00 | 0.00   | 0.03  | 0.01  | 0.09 | 0.87  |
| 1-2  | 0.70  | 0.21  | 0.01 | 0.00 | 0.00   | 0.03  | 0.03  | 0.12 | 1.12  |
| 1-3  | 0.95  | 0.25  | 0.01 | 0.37 | 0.50   | 0.50  | 0.09  | 0.18 | 2.87  |
| 1-4  | 1.59  | 0.50  | 0.03 | 0.00 | 0.00   | 0.18  | 0.20  | 1.98 | 4.50  |
| 1-5  | 0.50  | 0.17  | 0.00 | 0.00 | 0.00   | 0.03  | 0.03  | 0.09 | 0.82  |
| 1-6  | 1.14  | 0.40  | 0.01 | 1.78 | 1.81   | 2.21  | 0.89  | 0.29 | 8.56  |
| 1-7  | 0.60  | 0.20  | 0.01 | 0.00 | 0.00   | 0.07  | 0.03  | 0.10 | 1.04  |
| 1-8  | 0.78  | 0.28  | 0.01 | 0.18 | 0.18   | 0.12  | 0.03  | 0.29 | 1.90  |
| 1-9  | 1.54  | 0.59  | 0.03 | 1.00 | 1.03   | 1.37  | 0.48  | 1.56 | 7.62  |
| 1-10 | 0.29  | 0.10  | 0.00 | 0.00 | 0.00   | 0.01  | 0.00  | 0.04 | 0.46  |
| 2-1  | 0.50  | 0.15  | 0.00 | 0.00 | 0.00   | 0.01  | 0.01  | 0.07 | 0.76  |
| 2-2  | 0.92  | 0.40  | 0.03 | 1.50 | 1.21   | 2.15  | 3.29  | 4.64 | 14.17 |
| 2-3  | 0.73  | 0.26  | 0.01 | 0.00 | 0.00   | 0.06  | 0.06  | 0.14 | 1.28  |
| 2-4  | 0.48  | 0.20  | 0.00 | 0.09 | 0.06   | 0.04  | 0.01  | 0.07 | 0.98  |
| 2-5a | 0.45  | 0.18  | 0.01 | 0.00 | 0.00   | 0.06  | 0.03  | 0.07 | 0.82  |
| 2-5b | 0.15  | 0.06  | 0.00 | 0.01 | 0.03   | 0.01  | 0.00  | 0.04 | 0.32  |
| 2-6  | 0.84  | 0.29  | 0.01 | 0.00 | 0.00   | 0.14  | 0.09  | 0.15 | 1.54  |
| 2-7  | 1.01  | 0.40  | 0.03 | 0.85 | 0.73   | 0.81  | 0.18  | 0.34 | 4.39  |
| 2-8  | 0.50  | 0.21  | 0.01 | 0.00 | 0.00   | 0.07  | 0.04  | 0.14 | 1.00  |
| 2-9  | 1.20  | 0.37  | 0.01 | 0.96 | 1.17   | 1.35  | 0.14  | 0.31 | 5.54  |
| 2-10 | 0.62  | 0.25  | 0.00 | 0.00 | 0.00   | 0.04  | 0.01  | 0.07 | 1.01  |

> Build:   Building the disjuncts.
> Prune:   Pruning the disjuncts.
> Dup:     Eliminating duplicate disjuncts.
> And:     Constructing the *and* disjuncts (done only if *and* present).
> Prune2:  The second pruning phase (done only if *and* present).
> Power:   Power pruning.
> Count:   Counting the number of linkages (the exhaustive search).
> Post:    Post-processing all of the linkages.
> Total:   The total time of all phases.

1-1   Boris Yeltsin has demonstrated extraordinary courage in the defense of democratic values

1-2   yet his more recent claims of expanding his Russian republic 's lands understandably give his neighbors pause

1-3    like the Soviet Union , the republic is an artificial construct , embracing scores of nationalities and dozens of homelands

1-4    if Russian leaders can rally support for 12 million Russian people living in the Ukraine , Ukrainian leaders can bemoan the fate of six million of their compatriots in Russia

1-5    danger exists in having ethnic identity be too closely tied to republic borders

1-6    mechanisms must be developed to encourage the emergence of multiethnic societies and states , and to provide peaceful adjudication of border conflicts

1-7    despite his recent statements , Mr Yeltsin has handled the conflicts with considerable sensitivity , piecing together intricate republic-to-republic agreements

1-8    for their part , Russians have demonstrated surprisingly little bloodlust in light of their privations and dislocations

1-9    the victorious resistance to the coup is yet another manifestation of a commitment to democratic values that has been apparent in demonstrations , campaigns and kitchen table discussions for some time

1-10   a new Russian society has come into existence

2-1    the coup came at a bad time for Central Asia 's leaders

2-2    they must now demand independence or risk political defeat , despite the region 's overpopulation and underdevelopment that would be better helped by a more gradual transition

2-3    these artificial states were created by Soviet administrators to keep the region 's Muslim people from unifying

2-4    now the Central Asians must decide if they are a single people or five nationalities

2-5a   Uzbekistan is a viable independent state , but it would probably not be content with its present boundaries

2-5b   Uzbeks spill over into Kazakhstan , Kirghizia , Tadzhikistan and Turkmenistan

2-6    Kazakhstan is wealthy with natural resources , but it couldn't exist independently , as Kazakhs make up only 40 percent of the population

2-7    Russian people in Northern Kazakhstan would secede and seek annexation by Siberia , though this week 's tempest between Kazakhs and Yeltsin supporters was greatly overblown

2-8    if Kazakhstan splits apart , it will be hard for the other republics to resist Uzbek expansion

2-9    Uzbekistan is the cradle of the region 's Islamic civilization , and its leaders dream of a Central Asian empire

2-10   now this vision is secular , but deteriorating economies will favor Islamic radicalism

## D  The abridged dictionary

```
WALL: W+ or Q+;

John: (J- or O- or (S+ & {{@CO-} & {CL-}}) or SI-);
% All capitalized words - except at the beginning of sentences - will be
% treated as identical to "John".

dog cat man woman park yard bone neighbor store street bird hammer
nose party friend house movie brother sister diner student exam:
   {@A-} & ((Ds- & {@M+ or ({C+} & Bs+)} &
   (J- or Os- or (Ss+ & {{@CO-} & {CL-}}) or SIs-)) or Us-);

dogs cats men women parks yards bones neighbors stores streets birds
hammers noses parties friends houses movies brothers sisters diners
students exams wars winters actions laws successes:
   {@A-} & (({Dmc-} & {@M+ or ({C+} & Bp+)} & (J- or Op- or (Sp+ &
   {{@CO-} & {CL-}}) or SIp-)) or Up-);

water anger money politics trouble:
   {@A-} & (({Dmu-} & {@M+ or ({C+} & Bs+)} &
   (J- or Os- or (Ss+ & {{@CO-} & {CL-}}) or SIs-)) or Us-);

war winter action law success:
   {@A-} & (({D*u-} & {@M+ or ({C+} & Bs+)} &
   (J- or Os- or (Ss+ & {{@CO-} & {CL-}}) or SIs-)) or Us-);

she he: (Ss+ & {{@CO-} & {CL-}}) or SIs-;
me him them us: J- or O-;
her: D+ or J- or O-;
its my your their our: D+;
his: D+;

you: J- or O- or (Sp+ & {{@CO-} & {CL-}}) or SIp-;
it: J- or O- or (Ss+ & {{@CO-} & {CL-}}) or SIs-;
they we I: ((Sp+ & {{@CO-} & {CL-}}) or SIp-);

this: (J- or O- or (Ss+ & {{@CO-} & {CL-}}) or SIs-) or D*u+;
these: (J- or O- or (Sp+ & {{@CO-} & {CL-}}) or SIp-) or Dmc+;

something someone somebody everything nothing anything everyone
anyone everybody nobody anybody:
   {AI+ or @M+ or ({C+} & Bs+)} & (J- or O- or (Ss+ & {{@CO-} & {CL-}}) or SIs-);

those: (Dmc+) or (({P+} or {{C+} & Bp+}) &
       (J- or O- or (Sp+ & {{@CO-} & {CL-}}) or SIp- or Xb-));

the: D+;

both: Dmc+ or ({P+} & ((Sp+ & {{@CO-} & {CL-}}) or SIp- or J- or O- or Xb-));

biggest longest best worst favorite prettiest nicest same next
smartest stupidest ugliest shortest hardest easiest:  F-;
```

```
first second.a third last.a: F- or EV- or ({Xc+} & CO+);
other: Dm+;
an a every another: Ds+;

no: D+;

some: Dm+ or ({@M+ or ({C+} & Bp+)} & ((Sp+ & {{@CO-} & {CL-}}) or
   SIp- or J- or O-));
most: D+ or ({@M+ or ({C+} & B+)} & ((S+ & {{@CO-} & {CL-}}) or SI- or J- or O-));
2 two three four five six seven eight nine ten several: Dmc+ or
    ({@M+ or ({C+} & Bp+)} & ((Sp+ & {{@CO-} & {CL-}}) or SIp- or J- or O-));

many: Dmc+ or ({@M+ or ({C+} & Bp+)} & ((Sp+ & {{@CO-} & {CL-}}) or SIp- or J- or O-));
all: Dm+ or ({@M+ or J+ or ({C+} & B+)} & ((S+ & {{@CO-} & {CL-}}) or SI- or J- or O-));
one: Ds+ or ({@M+ or ({C+} & Bs+)} &
    ((Ss+ & {{@CO-} & {CL-}}) or SIs- or J- or O-));
any: D+ or ({@M+ or ({C+} & Bp+)} & ((Sp+ & {{@CO-} & {CL-}}) or SIp- or J- or O-));

did: (Q- & SI+ & I+) or ((S- or
    (Z- & B-)) & (((B- or O+) & {@EV+}) or I+));
do: (Q- & SIp+ & I+) or ((Sp- or
    (Z- & Bp-) or I- or W-) & (((B- or O+) & {@EV+}) or I+));
does: (Q- & SIs+ & I+) or ((Ss- or (Z- & Bs-)) & (((B- or O+) & {@EV+}) or I+));
done: (V- or M- or (T- & ((B- or O+) & {@EV+})));
doing: ((O+ & {@EV+} & ((G+ & {{@CO-} & {CL-}}) or GI- or M-)) or (GI- & B- & {@EV+}));

has: ((Q- & SIs+) or Ss- or (Z- & B-)) & (TO+ or ((B- or O+) & {@EV+}) or T+);
have: ((Q- & SIp+) or Sp- or (Z- & Bp-) or I- or W-) &
       (TO+ or ((B- or O+) & {@EV+}) or ({N+} & T+) or VC-);
had: ((Q- & SI+) or S- or (Z- & B-) or T-) & (TO+ or ((B- or O+) & {@EV+}) or T+);
having: (TO+ or ((B- or O+) & {@EV+}) or T+) & ((G+ & {{@CO-} & {CL-}}) or GI- or M-);

is was: ((G- or Ss- or (Z- & Bs-) or (Q- & (GI+ or SIs+)))
         & (AI+ or O+ or B- or P+ or GI+ or V+ or TO+ or TH+)) or (QI- & SIs+);
are were: ((Sp- or (Z- & Bp-) or (Q- & SIp+)) &
            (AI+ or P+ or O+ or B- or GI+ or V+ or TO+ or TH+)) or (QI- & SIs+);
be: (I- or W-) & (AI+ or IX- or P+ or B- or GI+ or V+ or TO+ or TH+);
been: T- & (AI+ or IX- or P+ or B- or GI+ or V+ or TO+ or TH+);
being: (AI+ or IX- or P+ or B- or GI+ or V+ or
        TO+ or TH+) & ((G+ & {{@CO-} & {CL-}}) or GI-);

will can.v may must could should would might:
    ((Q- & SI+) or S- or G- or (Z- & B-)) & I+;

run come: {@EX-} & (Sp- or (Z- & Bp-) or I- or W- or T-) & {@EV+};
runs comes goes: {@EX-} & (Ss- or (Z- & Bs-)) & {@EV+};
ran came went: {@EX-} & (S- or (Z- & B-)) & {@EV+};
go: {@EX-} & (Sp- or (Z- & Bp-) or I- or W-) & {@EV+};
gone: {@EX-} & T- & {@EV+};
going: {@EX-} & (GI- or M-) & {TO+} & {@EV+};
running coming: {@EX-} & (GI- or M-) & {@EV+};

talk arrive die:
    {@EX-} & (Sp- or (Z- & Bp-) or I- or W-) & {@EV+};
talks.v arrives dies:
```

```
    {@EX-} & (Ss- or (Z- & Bs-)) & {@EV+};
talked arrived died:
    {@EX-} & (S- or (Z- & B-) or T-) & {@EV+};
talking arriving dying:
    {@EX-} & (GI- or M-) & {@EV+};

move.v win lose fly:
    {@EX-} & (Sp- or (Z- & Bp-) or I- or W-) & {O+ or B-} & {@EV+};
moves.v wins loses flies:
    {@EX-} & (Ss- or (Z- & Bs-)) & {O+ or B-} & {@EV+};
moved won lost flew:
    {@EX-} & (V- or M- or ((S- or (Z- & B-) or T-) & {O+ or B-})) & {@EV+};
winning losing moving.v flying:
    {@EX-} & (GI- or M-) & {O+ or B-} & {@EV+};

meet destroy chase invite kick arrest:
    {@EX-} & (Sp- or (Z- & Bp-) or I- or W-) & (O+ or B-) & {@EV+};
hit: {@EX-} & (V- or M- or ((S- or (Z- & B-) or I- or W- or T-) &
    (O+ or B-))) & {@EV+};
meets destroys chases hits invites kicks arrests:
    {@EX-} & (Ss- or (Z- & Bs-)) & (O+ or B-) & {@EV+};
met destroyed chased invited kicked arrested:
    {@EX-} & (V- or M- or ((S- or (Z- & B-) or T-) & (B- or O+))) & {@EV+};
meeting.v destroying chasing hitting inviting kicking arresting:
    {@EX-} & (GI- or M-) & (O+ or B-) & {@EV+};

tell: {@EX-} & (Sp- or (Z- & Bp-) or I- or W-) & ((O+ or B-) &
    {TH+ or CLb+ or Zb+ or TT+ or R+ or @EV+});
tells: {@EX-} & (Ss- or (Z- & Bs-)) & ((O+ or B-) & {TH+ or CLb+ or Zb+ or
    TT+ or R+ or @EV+});
told: {@EX-} & (V- or M- or ((S- or (Z- & B-) or T-) & (O+ or B-))) &
    {TH+ or CLb+ or Zb+ or TT+ or R+ or @EV+};
telling: {@EX-} & (GI- or M-) & ((O+ or B-) & {TH+ or CLb+ or Zb+ or
    TT+ or @EV+});

ask: {@EX-} & (Sp- or (Z- & Bp-) or I- or W-) & ({O+ or B-} & {TT+ or R+ or @EV+});
asks: {@EX-} & (Ss- or (Z- & Bs-)) & ({O+ or B-} & {TT+ or R+ or @EV+});
asked: {@EX-} & (V- or M- or ((S- or (Z- & B-) or T-) & {O+ or B-})) &
    {TT+ or R+ or @EV+};
asking: {@EX-} & (GI- or M-) & ({O+ or B-} & {TT+ or R+ or @EV+});

wonder: {@EX-} & (Sp- or (Z- & Bp-) or I- or W-) & {R+ or @EV+};
wonders: {@EX-} & (Ss- or (Z- & Bs-)) & {R+ or @EV+};
wondered: {@EX-} & (S- or (Z- & B-) or T-) & {R+ or @EV+};
wondering: {@EX-} & (GI- or M-) & {R+ or @EV+};

want need.v: {@EX-} & (Sp- or (Z- & Bp-) or I- or W-) & (TO+ or ((O+ or B-) &
    {@EV+} & {TT+}));
wants needs.v: {@EX-} & (Ss- or (Z- & Bs-)) & (TO+ or ((O+ or B-) &
    {@EV+} & {TT+}));
wanted needed: {@EX-} & (((V- or M-) & {@EV+}) or ((S- or (Z- & B-) or T-)
    & (TO+ or ((O+ or B-) & {@EV+} & {TT+}))));
wanting needing: {@EX-} & (GI- or M-) & (((O+ or B-) & {@EV+} & {TT+}) or TO+);

demand.v: {@EX-} & (Sp- or (Z- & Bp-) or I- or W-) & (TO+ or ((O+ or B-)
```

```
    & {@EV+}) or TH+ or TS+ or (SI+ & I+));
demands.v: {@EX-} & (Ss- or (Z- & Bs-)) & (TO+ or
    ((O+ or B-) & {@EV+}) or TH+ or TS+ or (SI+ & I+));
demanded: {@EX-} & (((V- or M-) & {@EV+}) or ((S- or (Z- & B-) or T-)
    & (TO+ or ((O+ or B-) & {@EV+}) or TH+ or TS+ or (SI+ & I+))));
demanding: {@EX-} & (GI- or M-) & (TO+ or ((O+ or B-) & {@EV+}) or TH+ or TS+
    or (SI+ & I+));

start begin continue stop.v try:
    {@EX-} & (Sp- or (Z- & Bp-) or I- or W-) & {TO+ or ({O+ or B-} &
    {@EV+}) or GI+};
starts begins continues stops.v tries:
    {@EX-} & (Ss- or (Z- & Bs-)) & {TO+ or ({O+ or B-} &
    {@EV+}) or GI+};
started continued stopped tried: {@EX-} & (((V- or M-) & {@EV+})
    or ((S- or (Z- & B-) or T-) & {TO+ or ({O+ or B-} & {@EV+}) or GI+}));
began: {@EX-} & (S- or (Z- & B-) or T-) & {TO+ or ({O+ or B-} &
    {@EV+}) or GI+};
begun: {@EX-} & (V- or M-) & {@EV+};
starting beginning continuing stopping trying: {@EX-} & (GI- or M-) &
    (TO+ or ({O+ or B-} & {@EV+}) or GI+) & {@EV+};

see hear watch.v: {@EX-} & (Sp- or (Z- & Bp-) or I- or W-) & {(B- or O+) &
    {I+ or GI+}} & {@EV+};
sees hears watches.v: {@EX-} & (Ss- or (Z- & Bs-)) & {(B- or O+) &
    {I+ or GI+}} & {@EV+};
heard watched: {@EX-} & (V- or M- or ((S- or (Z- & B-) or T-) &
    {(B- or O+) & {I+ or GI+}})) & {@EV+};
saw: {@EX-} & (S- or (Z- & B-)) & (B- or O+) & {I+ or GI+} & {@EV+};
seen: {@EX-} & (V- or M- or (T- & (B- or O+) & {I+ or GI+})) & {@EV+};
seeing hearing watching: {@EX-} & (GI- or M-) & {{O+ or B-} & {I+ or GI+}} &
    {@EV+};

make: {@EX-} & (Sp- or (Z- & Bp-) or I- or W-) & ((B- or O+) & {@EV+} &
    {I+ or AI+});
makes: {@EX-} & (Ss- or (Z- & Bs-)) & ((B- or O+) & {I+ or AI+} & {@EV+});
made: {@EX-} & (V- or M- or ((S- or (Z- & B-) or T-) & (B- or O+))) &
    {I+ or AI+} & {@EV+};
making: {@EX-} & (GI- or M-) & ((O+ or B-) & {I+ or AI+}) & {@EV+};

think say:  {@EX-} & (Sp- or (Z- & Bp-) or I- or W-)
    & {CLb+ or TH+ or Z+};
thinks says: {@EX-} & Ss-
    & {CLb+ or TH+ or Z+};
thought said: {@EX-} & (V- or M- or
    ((S- or (Z- & B-) or T-) & {CLb+ or TH+ or Zb+}));
thinking saying: {@EX-} & (GI- or M-) &
    {CLb+ or TH+ or Zb+} & {@EV+};

expect claim.v: {@EX-} & (Sp- or (Z- & Bp-) or I- or W-) & (CLb+ or TH+ or
    Zb+ or TO+ or ((O+ or B-) & {@EV+} & {TT+}));
expects claims.v: {@EX-} & (Ss- or (Z- & Bs-)) & (CLb+ or TH+ or Zb+ or TO+ or
    ((O+ or B-) & {@EV+} & {TT+}));
expected claimed: {@EX-} & (((S- or (Z- & B-) or T-) & (CLb+ or TH+ or Zb+ or
    TO+ or ((O+ or B-) & {@EV+} & {TT+}))) or ((V- or M-) & {TO+}));
```

```
expecting claiming: {@EX-} & (M- or GI-) &
    (CLb+ or TH+ or Zb+ or TO+ or ((O+ or B-) & {@EV+} & {TT+})) & {@EV+};

give sell buy pass: {@EX-} & (Sp- or (Z- & Bp-) or I- or W-) & ((O+ or B-) &
    {O+} & {@EV+});
gives sells buys passes:
    {@EX-} & (Ss- or (Z- & Bs-)) & ((O+ or B-) & {O+} & {@EV+});
sold bought passed:
    {@EX-} & (V- or M- or ((S- or (Z- & B-) or T-) & (O+ or B-))) & {O+} & {@EV+};
gave: {@EX-} & (S- or (Z- & B-)) & (O+ or B-) & {O+} & {@EV+};
given: {@EX-} & (V- or M- or (T- & (O+ or B-))) & {O+} & {@EV+};
giving selling buying passing: {@EX-} & (GI- or M-) & (B- or O+) & {O+} & {@EV+};

look act.v sound.v: {@EX-} & (Sp- or (Z- & Bp-) or I- or W-) & AI+  & {@EV+};
looks acts.v sounds.v: {@EX-} & (Ss- or (Z- & Bs-)) & AI+ & {@EV+};
looked acted sounded: {@EX-} & (S- or (Z- & B-) or G- or T-) & AI+ & {@EV+};
looking acting sounding: {@EX-} & (GI- or M-) & AI+ & {@EV+};

ever hardly just probably also certainly partly largely never always
fortunately unfortunately apparently suddenly meanwhile eventually
then actually usually however moreover essentially commonly precisely
typically basically perhaps still presumably obviously potentially:
    EX+;

recently sometimes soon gradually specifically generally initially
ultimately already now sadly broadly:
    EX+ or EV-;

from with at against behind between below above during toward towards
without within beneath under beyond among for off in across through
by around out up down along like.p:
    (J+ or GI+) & (Mp- or EV-);

on over into about: (J+ or GI+ or R+) & (Mp- or EV-);

of: (J+ or GI+ or R+) & (Mp- or EV- or OF-);

here there: EV- or Mp-;

that: (CLb+ & TH-) or Ds+ or (C- & {Z+}) or SIs- or (Ss+ &
    {{@CO-} & {CL-}}) or J- or O- or (TS- & SI+ & I*j+) or J- or O-;

to: (I+ & (TO- or TV- or TT-)) or ((EV- or Mp-) & (J+ or GI+));

who: (C- & {Z+}) or QI+ or ((Q+ or R-) & B+) or (Ss+ & {R-});
what: ({U+} & (((Q+ or R-) & Br+) or ({R-} & Ss+) or QI+)) or
    ((Ss+ or Bs+) & (J- or O- or ({{@CO-} & {CL-}} & Ss+) or SIs-));
which: ({U+} & (((Q+ or R-) & Br+) or
    ({R-} & Ss+) or QI+)) or (C- & {Z+});

because unless though although but and: (CL+ & (({Xc+} & CO+) or EV-));
after before since until: (CL+ or GI+ or J+) & (({Xc+} & CO+) or EV- or Mp-);
if: CL+ & (({Xc+} & CO+) or EV- or R-);

when: (R- & (CL+ or IX+)) or Q+ or QI+ or (CL+ & (({Xc+} & CO+) or EV-));
```

```
why:   (R- & CL+) or Q+;
where:((R- or Mp-) & (CL+ or IX+)) or Q+ or QI+ or (CL+ & (({Xc+} & CO+) or EV-));
whether: R- & CL+;
how:   (R- & (CL+ or IX+)) or Q+ or QI+;

fast slow short long black white big small beautiful ugly tired angry:
   {Ea- or Eb+} & (A+ or (AI- & {EV+}));

glad afraid scared.a fortunate unfortunate lucky unlucky:
   {Ea-} & (A+ or (AI- & {EV+} & {CL+ or TO+ or TH+}));
certain uncertain sure unsure:
   {Ea-} & (A+ or (AI- & {EV+} & {CL+ or TO+ or TH+ or R+}));
happy correct.a incorrect sad right.a excited.a surprised.a delighted.a
disappointed.a upset.a sorry:
   {Ea-} & (A+ or (AI- & {EV+} & {TO+ or TH+}));
apprehensive secure optimistic pessimistic annoyed.a confused.a offended.a
insulted.a concerned.a confident depressed.a aware doubtful skeptical:
   {Ea- or Eb+} & (A+ or (AI- & {EV+} & {TH+}));
smart intelligent wise eager reluctant able unable impatient eligible:
   {Ea- or Eb+} & (A+ or (AI- & {EV+} & {TO+}));

almost nearly fairly pretty.e very quite extremely: Ea+;

quickly angrily naively: EV- or EX+;

too: ((AI- & AI+) or (EV- & EV+)) & {TO+};

so: (AI- & AI+ & {TH+}) or (EV- & CL+);

as: (CL+ or J+) & (({Xc+} & CO+) or Mp- or EV-);

more: Dm+ or (AI- & AI+ & {LM+}) or (O- & {U+} & {LN+}) or
   (EV- & {EV+} & {LE+});

bigger taller older younger heavier lighter darker fatter thinner
cheaper prettier uglier smaller larger deeper longer shorter stronger
weaker:
   {Ec-} & ((AI- & {EV+} & {LA+}) or A+);

than: ((IX+ or O+) & (LM- or LA-)) or ((LN- or LE-) & (O+ or IX+))
   or (LM- & AI+);

,: Xc-;
```

## E  The context-freeness of link grammars

We shall use the terms *word*, *terminal* or *terminal symbol* interchangeably. We shall use the terms *non-terminal*, *non-terminal symbol* and *variable* interchangeably.

**Theorem 1:** *The class of languages that can be expressed as basic link grammars (as defined in section 2.1) is the class of context-free languages.*

**Remark:** Our proof (below) that a link grammar is context free is somewhat cumbersome. There is a simpler way in which to gain intuition about this theorem.[14] It is a well known theorem that the languages accepted by a non-deterministic push-down automaton (NDPA) are the class of context-free languages [5].

Consider an NPDA that processes a sequence of words from left to right. Each element of the stack is a connector name. When the NPDA encounters a word, it chooses one of the disjuncts of that word (non-deterministically) that has the property that its left-pointing connectors match the connectors on the top of the stack (the nearest connecting connector matching the top of the stack, etc.). It pops these connectors off of the stack, and pushes the right pointing connectors of that disjunct onto the stack (the far matching connector gets pushed first, etc.). If the sequence of words is a sentence of the link grammar, then there exists a way for this NPDA to start with an empty stack, process all the words, and end with an empty stack. This follows immediately from the linkage. The contents of the stack after processing a word $W$ (but before processing $W + 1$) consists of the links that cross the vertical line drawn between $W$ and $W + 1$. (The lowest link corresponds to the top of the stack, etc.)

The reason that this is not a proof of the theorem is that this NPDA also accepts sequences of words that are not in the link grammar. For example, a linkage which is not connected also leads to an accepting computation for this NPDA. The exclusivity condition is also difficult to enforce. Rather than fixing this proof we give a different proof (below) that generates an appropriate context-free grammar directly from the link grammar. $\square$

*Proof:*  First we show that the language of any context-free grammar can be expressed as a link grammar, then we show the converse.

Given any context-free language $L$, there is a context-free grammar $G$ for $L$. We may assume without loss of generality that $G$ is expressed in Greibach normal form [5, p. 95]. This means that every production is of the form
$$A \to xA_1A_2 \cdots A_k$$
where $A, A_1, \ldots, A_k$ are all variables (or non-terminal symbols) and $x$ is a terminal symbol. The number of variables on the right side, $k$, may be zero. We may also assume without loss of generality that the start non-terminal symbol occurs on the left side of exactly one production, and on the right side of none.

The link grammar we construct will have the same terminal symbols as $G$. For each production in which a terminal symbol $x$ appears, one disjunct is included in the the set of disjuncts for $x$. In particular, for the production shown above (in which $A$ is not the start non-terminal symbol), the disjunct generated is:

$$((\text{A}) \ (\text{A}k,\ldots,\text{A}1))$$

---

[14]Thanks to Avrim Blum and Gary Miller for pointing this out.

If $A$ is the start non-terminal symbol for the grammar, then the disjunct for $x$ is:

$$(() \ (\text{A}k,\ldots,\text{A}1))$$

Note that the connector names in the link grammar correspond to variables in the context free grammar.

We claim that the language accepted by this link grammar, $LG$, is the same as the language of the context-free grammar $G$. To justify this claim, we first show that any sequence of terminals $X$ accepted by $G$ is also accepted by $LG$. Consider the derivation tree of $X$ in the grammar $G$. Each internal node of the tree corresponds to a variable, and the leftmost child of each such node is a terminal symbol. If we merge each variable with its leftmost child (a terminal symbol), then the remaining edges of the derivation tree are in one-to-one correspondence with a set of links which satisfy all the conditions required to show that $X$ is in $LG$.

It remains to show that any sequence of words accepted by $LG$ is also accepted by $G$. Consider a way to apply links to a string $X$ which satisfy the requirements of $LG$. Define the parent of a word in the sequence $X$ to be the word to which it is connected through a link to its left (if there is such a link). The parent of each word is unique, since there is at most one left-pointing connector in each disjunct. Every link is accounted for in this manner. This shows that the links and words form a forest of trees. Since the links must connect the words together, they must form a single tree. This tree is isomorphic to a merged derivation tree (mentioned above) for the string $X$. This completes the first half of the proof of Theorem 1.

The proof that for every link grammar there is a context free grammar that accepts the same language is more involved than the previous one. Here we take a link grammar (in disjunctive form) and construct a context-free grammar that accepts the same language. The variables of our context-free grammar will correspond to cross products of initial and final subsequence of connector names occurring in the disjuncts in the link grammar. To be more specific, consider the following two disjuncts that occur in the link grammar:

$$((L_1,\ldots,L_m) \ (R_n,\ldots,R_1)) \qquad ((L'_1,\ldots,L'_{m'}) \ (R'_{n'},\ldots,R'_1))$$

Then for every $0 \leq i \leq n$ and every $0 \leq j \leq m'$ there is a variable in our context-free grammar $V_\alpha$, where

$$\alpha \ = \ ((R_i,\ldots,R_1) \ (L'_1,L'_2,\ldots,L'_j))$$

A variable in a context-free grammar is ultimately (in any derivation) replaced by a sequence of terminal symbols. Before we specify the productions of our context-free grammar, it is instructive to describe the terminal sequences which the non-terminals will ultimately expand into.

Let $\alpha$ be defined as above, and let $V_\alpha$ be a variable. This variable can lead to a sequence of words $X$ if an only if $X$ has the following property:

> The sequence $aXb$ is in the given link grammar modified by adding two additional words $a$ and $b$, where the single disjunct on $a$ is
>
> $$(() \ (Q, R_j, \ldots, R_1))$$
>
> and the single disjunct on $b$ is
>
> $$((L'_1, L'_2, \ldots, L'_j, Q) \ ()).$$
>
> $Q$ is a connector name different from all others. This requires that there is a link from the $Q$ connector on $a$ to that on $b$.

Alternatively, consider a subsequence of consecutive words in a sentence of the link grammar that has the property that

(1) When it is deleted, the rest of the sentence remains connected together.

(2) It connects to the rest of the sentence through the words neighboring either end of the sequence.

(3) After deleting the subsequence of words, the unresolved connectors on the left side of the gap are $(R_j, \ldots, R_1)$ and the unresolved connectors on the right side of the gap are $(L'_1, L'_2, \ldots, L'_j)$.

Then any sequence derivable from $V_\alpha$ can replace the deleted subsequence and still give a sentence of the link grammar.

There are three types of productions: start productions, the epsilon production, and general productions. We now describe these in detail.

**Start Productions:** For every word $x$ with a disjunct of the form:

$$((L_1, L_2, \ldots, L_m) \ ())$$

generate a production
$$S \to V_\alpha x$$
where $\alpha = (() \ (L_1, L_2, \ldots, L_m))$ and $S$ is the start variable of the context-free grammar.

**The epsilon production:**
$$V_\alpha \to \epsilon$$
where $\alpha = $ (() ()) and $\epsilon$ is the empty string.

**General productions:** A general production has the form
$$V_\alpha \to V_\beta \ x \ V_\gamma$$
where subscripts $\beta$ and $\gamma$ are chosen as a function of $\alpha$ and one of the disjuncts on the word $x$. Let
$$\alpha = ((R_n, \ldots, R_1)\ (L_1, \ldots, L_m))$$
and let one of the disjuncts on $x$ be:
$$((L'_1, \ldots, L'_{m'})\ (R'_{n'}, \ldots, R'_1))$$

(Note that any of $n$, $m$, $n'$ and $m'$ could be zero.) For a particular choice of $\alpha$ and a disjunct on $x$ up to three productions could be generated.

**(a)** The connector $L'_{m'}$ matches $R_n$ and $m', n \geq 1$.

$$\begin{aligned}\beta &= ((R_{n-1}, \ldots, R_1)\ (L'_1, \ldots, L'_{m'-1})) \\ \gamma &= ((R'_{n'}, \ldots, R'_1)\ (L_1, \ldots, L_m))\end{aligned}$$

**(b)** The connector $R'_{n'}$ matches $L_m$ and $m, n' \geq 1$.

$$\begin{aligned}\beta &= ((R_n, \ldots, R_1)\ (L'_1, \ldots, L'_{m'})) \\ \gamma &= ((R'_{n'-1}, \ldots, R'_1)\ (L_1, \ldots, L_{m-1}))\end{aligned}$$

**(c)** The connector $L'_{m'}$ matches $R_n$, $R'_{n'}$ matches $L_m$ and $m, m', n, n' \geq 1$.

$$\begin{aligned}\beta &= ((R_{n-1}, \ldots, R_1)\ (L'_1, \ldots, L'_{m'-1})) \\ \gamma &= ((R'_{n'-1}, \ldots, R'_1)\ (L_1, \ldots, L_{m-1}))\end{aligned}$$

Given a derivation of a string $X$, we can construct a linkage for showing $X$ is in the language of the link grammar. Given a linkage of a sentence $X$, we can construct a derivation that shows that $X$ is in the language of the context free grammar. It follows that the two grammars accept the same language. We now explain these two constructions.

In any derivation, we can postpone application of the epsilon productions until the very end. Any intermediate state of the derivation therefore has the property that it consists of alternating variables and words. Whenever a general production of type (a) is used, generate a link between the new word $x$ and the word to right of variable $V_\gamma$. Whenever a general production of type (b) is used, generate a link between $x$ and the word to left of variable $V_\beta$. Whenever a general production of type (c) is used, generate links between $x$ and the word to left of variable $V_\beta$ and the word to the right of variable $V_\gamma$. At any point in the derivation, all of the words generated so far have been linked together. Furthermore, the subscripts of the variables that are left correspond to connectors that have yet to be satisfied by links. At the end, the fact that the epsilon productions can be applied to rid the sequence of variables, guarantees that all of the connectors on the chosen disjuncts have been satisfied.

To transform a linkage into a derivation, we will proceed from top to bottom. To get started we apply the start production corresponding to the disjunct that is used in the rightmost word of the sentence. The non-terminal resulting from this application will eventually expand into the whole sentence minus the last word. In the description that follows, we'll maintain the induction that every non-terminal will eventually expand into a specific sequence of terminals satisfying conditions (1), (2) and (3) (from before the proof).

To decide which production to apply to a given non-terminal $V_\alpha$, we consider the range $[L, R]$ of non-terminals $V_\alpha$ is supposed to expand into. Either word $L-1$ or word $R+1$ must be connected to some word in this range. Let $x$ be the rightmost word to which $L$ is connected (or the leftmost word to which $R$ is connected). If $x$ is connected to $L-1$ and not $R+1$, then we apply a general production of type (a). If $x$ is connected to $R+1$ and not to $L-1$, then we apply a general production of type (b). If it is connected to both, then we apply a general production of type (c). The ranges of the two resulting non-terminals become the original range minus $x$ and everything to its right, and the original range minus $x$ and everything to its left. If the range is empty, then the epsilon production can be applied. This construction is analogous to the way in which the algorithm of section 4 finds a linkage. □

Corollary: *The languages accepted by a link grammar [with or without multi-connectors], [with or without the exclusion rule], and [with or without arbitrary connector matching] are all context-free.*

*Proof:* Recall that a multi-connector is a connector that can attach to more than one link, the exclusion rule requires that no two links connect the same pair of words, and an arbitrary matching allows the link grammar to specify an arbitrary table specifying which connector pairs match. It is easy to modify the proof of theorem 1 to handle all of these cases. The only change is in the general productions. We omit the details of this proof. □